\documentclass[lettersize,journal]{IEEEtran}
\usepackage{amsfonts}
\usepackage{algorithmic}
\usepackage{algorithm}
\usepackage{array}
\usepackage[caption=false,font=normalsize,labelfont=sf,textfont=sf]{subfig}
\usepackage{textcomp}
\usepackage{stfloats}
\usepackage{url}
\usepackage{verbatim}
\usepackage{graphicx}
\usepackage{cite}
\usepackage{graphicx}
\usepackage{amsmath}
\usepackage{amssymb}
\usepackage{multirow}
\usepackage{subfiles}
\usepackage{pgfplots}
\usepackage{makecell}
\usepackage[colorlinks,linkcolor=red]{hyperref}
\usepackage{bm}
\usepackage{marvosym}

\newcommand{\SOTA}{state-of-the-art}
\newcommand{\SOTABIG}{State-of-the-Art}
\newcommand{\ETHUCY}{ETH-UCY}
\newcommand{\NA}{NA}

\newcommand{\MARK}[1]{\textcolor{blue}{\textbf{#1}}}
\newcommand{\MARKR}[1]{\textcolor{purple}{\textbf{#1}}}

\newcommand{\SOCIALGAN}{Social GAN}
\newcommand{\SOPHIE}{SoPhie}
\newcommand{\BIGAT}{Social-BiGAT}
\newcommand{\STGAT}{STGAT}
\newcommand{\STGCNN}{Social-STGCNN}
\newcommand{\SIMAUG}{SimAug}
\newcommand{\PECNET}{PECNet}
\newcommand{\SRLSTM}{SR-LSTM}
\newcommand{\SRPAMI}{E-SR-LSTM}
\newcommand{\GARDEN}{Multiverse}
\newcommand{\PEEKING}{Next}
\newcommand{\TRANSFORMER}{STAR}
\newcommand{\TPNMS}{TPNMS}
\newcommand{\TRAJECTRONPP}{Trajectron++}
\newcommand{\LBEBM}{LB-EBM}

\newcommand{\TF}{TF}
\newcommand{\PCCS}{PCCSNet}
\newcommand{\STC}{STCNet}
\newcommand{\YNET}{$\textsf{Y}$-net}
\newcommand{\UNIN}{UNIN}
\newcommand{\SSALVM}{SSALVM}
\newcommand{\SPECTGNN}{SpecTGNN}
\newcommand{\CSC}{CSCNet}
\newcommand{\MID}{MID}
\newcommand{\MEMO}{MemoNet}
\newcommand{\VDRGCN}{VDRGCN}
\newcommand{\SHENET}{SHENet}
\newcommand{\SEEM}{SEEM}
\newcommand{\SSL}{Soacil-SSL}
\newcommand{\NSP}{NSP-SFM}

\newcommand{\MODEL}{MSN}
\newcommand{\MODELD}{MSN-D}
\newcommand{\MODELG}{MSN}

\newcommand{\FIG}[1]{\figurename~\ref{#1}}
\newcommand{\TABLE}[1]{TABLE \ref{#1}}
\newcommand{\EQUA}[1]{Equation \ref{#1}}
\newcommand{\ETAL}{{\textit{et al.}}}

\newcommand{\IE}{\emph{i.e.}}
\newcommand{\EG}{\emph{e.g.}}
\newcommand{\TO}{$\to$~}

\newcommand{\DAGGER}{$^\dagger$}

\begin{document}

\title{MSN: Multi-Style Network for Trajectory Prediction}

\author{
    Conghao Wong*,
    Beihao Xia*,
    Qinmu Peng,
    Wei Yuan,
    and Xinge You (\Letter),~\IEEEmembership{Senior Member,~IEEE}
\thanks{
    Authors are with Huazhong University of Science and Technology, Wuhan, Hubei, P.R.China.
    Email: conghaowong@icloud.com, xbh\_hust@icloud.com, pengqinmu@hust.edu.cn, yuanwei@hust.edu.cn, youxg@hust.edu.cn.
}
\thanks{
    Codes are available at \url{https://github.com/NorthOcean/MSN}.
}
\thanks{
    ``*'' indicates authors with equal contribution.\\
}
}

\markboth{Journal of \LaTeX\ Class Files,~Vol.~14, No.~8, August~2021}%
{Shell \MakeLowercase{\textit{et al.}}: A Sample Article Using IEEEtran.cls for IEEE Journals}


\maketitle

\begin{abstract}
Trajectory prediction aims to forecast agents' possible future locations considering their observations along with the video context.
It is strongly needed by many autonomous platforms like tracking, detection, robot navigation, and self-driving cars.
Whether it is agents' internal personality factors, interactive behaviors with the neighborhood, or the influence of surroundings, they all impact agents' future planning.
However, many previous methods model and predict agents' behaviors with the same strategy or feature distribution, making them challenging to make predictions with sufficient style differences.
This paper proposes the Multi-Style Network (MSN), which utilizes style proposal and stylized prediction using two sub-networks, to provide multi-style predictions in a novel categorical way adaptively.
The proposed network contains a series of style channels, and each channel is bound to a unique and specific behavior style.
We use agents' end-point plannings and their interaction context as the basis for the behavior classification, so as to adaptively learn multiple diverse behavior styles through these channels.
Then, we assume that the target agents may plan their future behaviors according to each of these categorized styles, thus utilizing different style channels to make predictions with significant style differences in parallel.
Experiments show that the proposed MSN outperforms current state-of-the-art methods up to 10\% quantitatively on two widely used datasets, and presents better multi-style characteristics qualitatively.
\end{abstract}

\begin{IEEEkeywords}
    Trajectory prediction, Multi-style, Hidden behavior category, Transformer.
\end{IEEEkeywords}

\section{Introduction}
\label{Introduction}

\begin{figure}[tbp]
    \centering
    \includegraphics[width=0.5\textwidth]{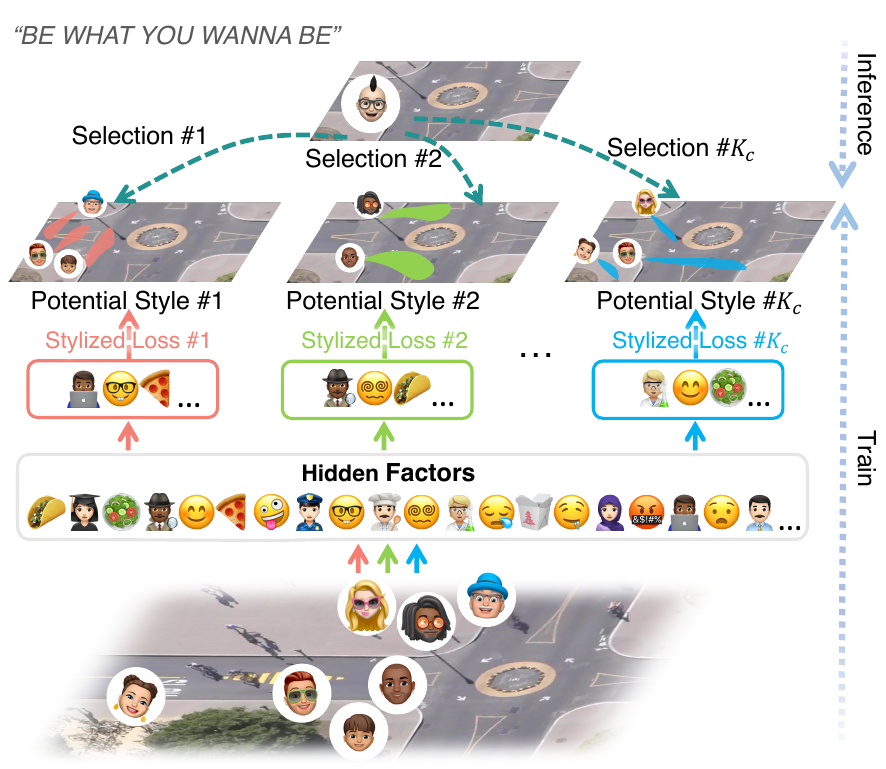}
    \caption{
        Agents' personalities and preferences may be affected by a large number of hidden factors, thus presenting a variety of potential behavior styles.
        When we do not know much about the target agents or their behavior preferences, a natural thought is to sequentially assume that they behave with the specific behavior styles.
        Then, we could predict all potential future choices under these assumptions, thus giving a series of future predictions with significant style differences.
        Notably, all potential trajectories will not be ranked.
        Agents are encouraged to \emph{be what they wanna be}.
        We call this strategy the \emph{Multi-Style Prediction}.
    }
    \label{fig_intro}
\end{figure}

\IEEEPARstart{A}{}{NALYZING} and understanding agents' activities or behaviors is beneficial and crucial, but challenging in the intelligent world.
This paper explores agents' behaviors through trajectory prediction, a novel but practical task.
Trajectory prediction forecasts agents' potential future trajectories based on their historical locations, surroundings, and possible interactive behaviors.
Its application is becoming increasingly widespread, for trajectories are always easier to obtain and more quickly to analyze than other descriptions with different modalities.
The analyzed results within this task and its forecasted trajectories could serve a lot of intelligent systems and platforms, such as object detection \cite{mehran2009abnormal,2016Anomaly}, object tracking \cite{youWillNeverWalkAlone,saleh2020artist}, robotic navigation \cite{unfreezing}, autonomous driving \cite{deo2018convolutional,deo2018would,cui2019multimodal,li2019pedestrian,8658195}, behavior analysis \cite{alahi2017learning,chai2019multipath,Phan-Minh_2020_CVPR}.
However, whether it is the interaction between heterogeneous agents, the inhomogeneous physical environment, or the predicted agents' uncertain intentions and different behavioral styles, all these factors will pose significant challenges in the prediction process.
In addition, agents' varying response preferences and independent awareness also make this task more difficult.

\textbf{Agents' Behavior Styles.}
Previous researchers have investigated several factors that influence agents' future plannings, including the social interaction (the agent-agent interaction) \cite{socialLSTM,xia2020bgm,9043898} and the scene interaction (agent-scene interaction) \cite{sophie,bigat}.
Most methods make socially and physically acceptable predictions by considering these two interactive factors together.
Additionally, other researchers \cite{socialGAN,sophie,amirian2019social} have extensively studied stochastic ways to reflect agents' uncertain future choices and multi-path preferences.
However, most existing approaches forecast agents' possible future choices with the same ``style'', even for most current stochastic methods.

The word ``style'' comes from image-related tasks.
For example, \cite{lin2017bilinear} models images as the two-factor variations such as ``style'' and ``content''.
Other researchers have also studied transferring images' styles into another, known as the \emph{style transfer}.
Like the ``style'' mentioned in images \cite{karras2019style}, agents' behaviors and trajectories also present different ``styles''.
The styles could reflect different preferences for making decisions and trajectory plans.
We call the styles presented by trajectories the \emph{Behavior Styles}.
For even agents with almost identical historical states, their future choices could be vastly different due to their different behavior styles.
For example, a father and his son go out in the morning.
The son may turn at the crossroads to go to school like a ``student'', while the father may head to the office for he is an ``employee''.
When making predictions about someone we do not know a lot, we can provide predictions with the above two distinct styles, including both ``student'' and ``employee'', so as to achieve the multi-style trajectory prediction goal.


\begin{figure*}[tbp]
    \centering
    \includegraphics[width=0.8\textwidth]{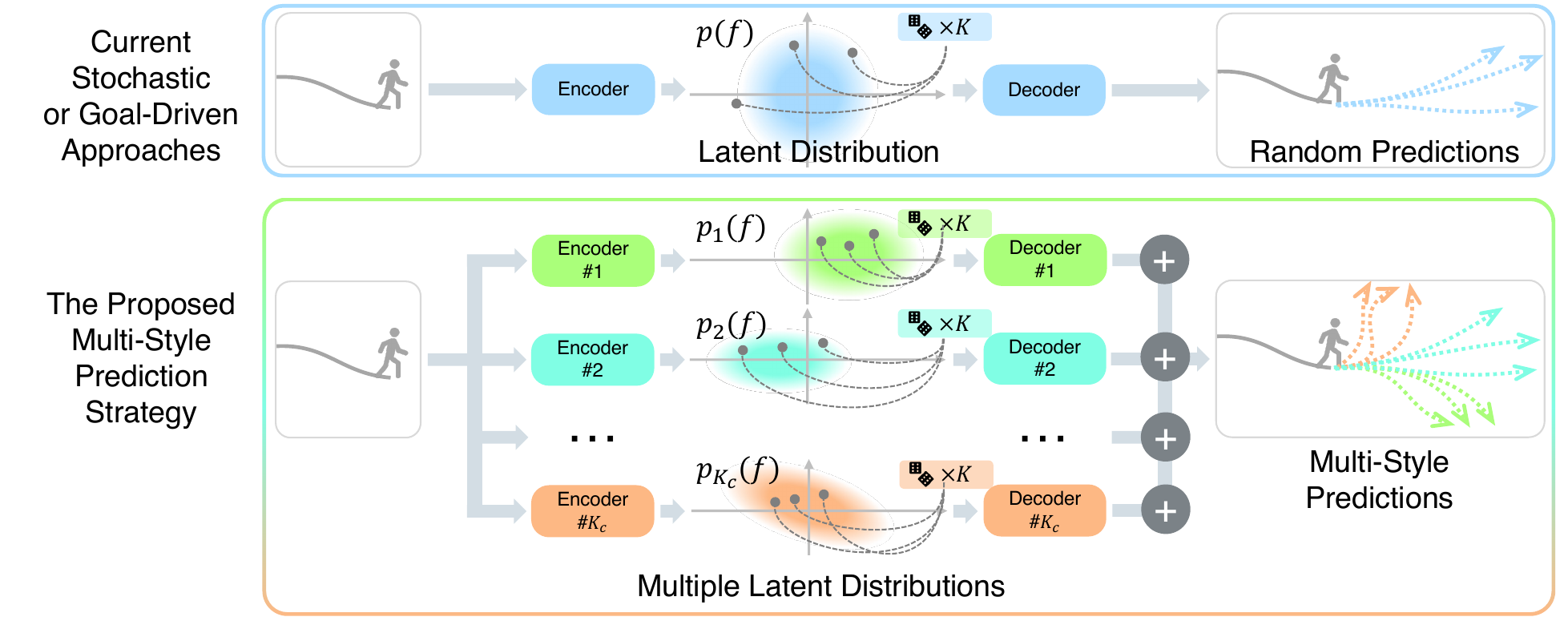}
    \caption{
        Illustration of the difference between multi-style prediction strategy and previous approaches.
        Most current stochastic or goal-driven methods encode agents' behavior representations into one latent distribution $p(f)$, and them randomly sample features to obtain multiple stochastic predictions. 
        The proposed multi-style prediction strategy employs $K_c$ encoder-decoder pairs, and each of them learns agents' latent representations $p_{k}(f)$ according to their categorized behavior styles.
        Then, we sample features in each latent distribution in parallel, thus providing multi-style predictions with sufficient style variation.
    }
    \label{fig_intro_style}
\end{figure*}

Stochastic trajectory prediction \cite{desire,amirian2019social,chai2019multipath} have been widely studied.
They could forecast multiple random paths.
However, it is still challenging for them to describe and analyze such vastly future choices through random factors or just sampling features from the only latent representations.
In other words, they mostly encode agents' behavior representations in a ``generalized'' way.
In terms of occupation, educational experience, or personality, all have different impacts on agents' trajectory plans.
Furthermore, there is always something difficult to describe by specific indicators or quantitative metrics that plays an essential role in influencing agents' plans \emph{behind the prediction scenes}.
These factors directly determine agents' behaviors and plans, yet it is not easy to specify them directly with equations.
Whichever factor changes, agents' plans may differ, leading to multiple behavior styles.

\textbf{Classification for Behavior Styles.}
Agents' trajectories and behavioral preferences are distributed continuously.
It is challenging to determine behavior styles quantitatively.
A natural thought is to establish connections between trajectories and these stylized factors, thereby reflecting the potential impact of these invisible factors on their trajectories \emph{categorically}, \IE, classify agents' behaviors into several categories that could reflect the co-influence of hidden factors on trajectories.

When agents plan their short-term trajectories, they may make a general proposal for a specific destination and then take detailed actions purposefully according to the chosen destination.
Thus, different destination choices may reflect their different behavioral styles.
It is worth noting that some approaches, like \cite{rhinehart2019precog,mangalam2020not,mangalam2020s,tran2021goal}, have attempted to conduct \emph{goal-driven} trajectory prediction via agents' destinations, and they have achieved excellent prediction performance.

Inspired by these approaches, we use the similarities between agents' destinations to classify their behavior styles adaptively.
We call these categories the \textbf{Hidden Behavior Categories} since we do not know the distribution and presentation of trajectories for each category.
Thanks to these categories, we sequentially assume that the target agent may behave like each one of all these categories when making predictions.
Then, we employ the corresponding trajectory generator to forecast under the corresponding style assumptions, thus obtaining a series of \emph{Multi-Style} predictions.
In short, we attempt to use a series of generators to learn to generate multiple latent representations corresponding to trajectories with different behavior categories separately, thus giving stylized predictions.
As shown in \FIG{fig_intro}, we call the above prediction strategy the \emph{Multi-Style Prediction}.

\textbf{Multi-Style v.s. Single-Style Prediction.}
Some researchers have employed generative models, like Generative Adversarial Networks (GANs)\cite{socialGAN,sophie} and Variational Autoencoders (VAEs)\cite{trajectron}, to study agents' uncertainty and randomness, thus forecasting multiple future paths.
These methods may encode agents' trajectories and interactive behaviors into a specific data distribution (like the normalized Gaussian distribution), and then sample features to generate multiple trajectories during each model implementation.
However, agents' style attributes may be compressed when encoding their behaviors into the only ``uncategorized'' latent space.
Therefore, these generated trajectories would lose their corresponding stylized semantics.
Some methods \cite{deo2018would,kothari2020human} have further modeled interactions by manually labeling trajectory classifications.
However, they may consume many labeling costs and be challenging to adapt to a broader range of prediction scenarios.

Although existing goal-driven methods \cite{mangalam2020not,mangalam2020s} have achieved significant progress, their prediction strategies do not match the expectant multi-styled prediction.
The main difference is that multi-style prediction strategies classify agents into different categories through their destinations.
Then, generators in each category will learn the latent spaces with the corresponding style separately, thus making predictions for each behavioral category.
In this process, the destination plays an intermediate factor in judging their behavior categories, which are not yet considered in these approaches.

Different from current stochastic and goal-driven methods (called the \emph{single-style} prediction methods), the proposed multi-style prediction method could provide multi-style predictions for the same agent with the help of the above \emph{Hidden Behavior Categories} and the corresponding classification strategy.
Different trajectory generators perform their duties within each category to forecast trajectories with different styles separately, thus avoiding the above problems and then adaptively predicting trajectories with categorical differences simultaneously.
Details are shown in \FIG{fig_intro_style}.

This paper proposes the multi-style network \MODEL~to explore the stylized prediction.
In detail, \MODEL~ first classifies trajectories along with their interactive video context into several hidden behavior categories according to their destination choices adaptively.
Then, it employs and trains a generator with a series of style channels to learn how agents' stylized behavior representations distribute in each style, and then make multiple stylized planning proposals.
Finally, the model implements the stylized predictions based on the given proposals in each style channel by considering the potential interactive behaviors.
Our contributions are summarized as follows:

\begin{itemize}
    \item We propose a \emph{multi-style} prediction strategy by employing an adaptive classification strategy to classify agents' behaviors into multiple behavior categories.
    \item We design the \MODEL, which contains \emph{style proposal} and \emph{stylized prediction} two sub-networks, to forecast multi-style predictions for agents along with the novel behavior-category-based \emph{stylized loss}.
    \item Experiments demonstrate that \MODEL~outperforms current \SOTA~methods on \ETHUCY~and SDD, and show better multi-style characteristics.
\end{itemize}

\section{Related Work}

This paper focuses on categorically modeling agents' behaviors and predicting their trajectories in a novel multi-style way.
We summarize relevant approaches, including Context-Based Trajectory Prediction, Stochastic and Goal-Driven Trajectory Prediction (``Single-Style Trajectory Prediction''), and Trajectory Categorization Approaches.

\subsection{Context-Based Trajectory Prediction}
Previous works \cite{ssLSTM,rudenko2020human,quan2021holistic} have studied trajectory prediction by considering scene constraints and agents' potential interaction behaviors.
Alahi \ETAL~\cite{socialLSTM} treat this task as the sequence-data generation problem and adopt the Long Short-Term Memory (LSTM) to model agents' behaviors.
They also consider social interaction through a social pooling mechanism, which could share cell states among the target agent and its potential interactive neighbors.
After that, methods like agent-wise pooling \cite{socialGAN}, sorting \cite{sophie,bigat}, agent-aware attention mechanisms \cite{socialAttention,cidnn,srLSTM,zhang2020social} are proposed to further model interactive behaviors among agents.
In addition to social interaction, some researchers have studied how physical constraints and scene objects affect agents' future plannings, named scene interaction or agent-object interaction.
Methods like \cite{ssLSTM,lisotto2019social,manh2018scene} utilize the Convolutional Neural Networks (CNNs) to extract scene semantics from RGB images or videos, and therefore help establish connections between trajectories and images, which further helps predict physically acceptable predictions.
Moreover, \cite{peekingIntoTheFuture,liang2020temporal,liang2020simaug} adopt scenes' segmentation maps to model how different kinds of physical components or objects affect agents' decisions.
Researchers also employ Graph Attention Networks (GATs) \cite{socialAttention,stgat} and Graph Convolution Networks (GCNs) \cite{stgcnn} as their backbones to model interactions with the graph structures to obtain better interactive representations.
Transformers \cite{yu2020spatio,giuliari2020transformer} are also employed to obtain better interactive representations.

However, most of these methods do not consider the diverse interactions, let alone the multi-style characteristics of behaviors.
Furthermore, it is difficult for them to fully cover the different preferences presented in the trajectories.

\subsection{Single-Style Trajectory Prediction}
Recently, researchers have focused more on modeling the multimodal characteristics of agents' future choices.
The single-style trajectory prediction mentioned in this paper includes stochastic and goal-driven trajectory prediction methods.
These methods all aim at modeling agents' multiple future choices.
Generative neural networks, like VAEs \cite{desire,trajectron} and GANs \cite{socialGAN,amirian2019social,sophie,bigat}, are widely used to generate multiple potential trajectories through randomly sampling features from the latent feature space.
Compared with previous single-output prediction methods (also named the deterministic models), stochastic methods could better reflect agents' diverse interactive preferences and uncertain future choices, for they could make different predictions per implementation.

Some researchers have attempted to attribute agents' uncertain future choices to the uncertainty in their intentions or destination plannings.
Researchers like \cite{rehder2015goal,rehder2018pedestrian,mangalam2020not,mangalam2020s,rhinehart2019precog,rhinehart2018r2p2} have brought agents' destination choices to the trajectory prediction task to explore their multiple trajectory plannings further.
Rehder \ETAL~\cite{rehder2015goal} propose a destination-conditioned method to predict trajectories.
After that, Rehder \ETAL~\cite{rehder2018pedestrian} utilize a deep learning approach to simultaneously handle intention recognition and trajectory prediction.
Mangalam \ETAL~\cite{mangalam2020not} propose the endpoint-conditioned trajectory prediction method, and try to address agents' random endpoint choices by gathering scene segmentation maps and agents' trajectory waypoints in \cite{mangalam2020s}.
Quan \ETAL~\cite{quan2021holistic} employ an extra intention LSTM cell to capture the intention cues and their changing trends.
Moreover, some methods use scene context like scene images \cite{tran2021goal,quan2021holistic} to infer agents' possible goal choices, therefore helping forecast multiple stochastic predictions.

However, these stochastic or goal-driven methods, which we call together the \emph{single-style} methods, forecast multiple predictions through randomly sampling features from the ``only'' distribution for all agents, resulting in a ``single'' prediction style.
Besides, most current goal-driven approaches barely consider the diversity of agents' behavior styles, making them difficult to make predictions with sufficient variability.
To address this problem, some researchers have tried to improve and use specific sampling methods, like the Truncation Trick\cite{mangalam2020not} and the Test-Time Sampling Trick\cite{mangalam2020s}, to sample features from the latent distribution, thus obtaining predictions with better multimodal characteristics.
Furthermore, Chen \ETAL~\cite{chen2021personalized} propose a distribution discrimination method to learn the latent distribution to represent agents' different motion patterns, and optimize it by the contrastive discrimination.

Furthermore, using a ``single'' distribution to describe the diverse and differentiated behaviors may not avoid compressing their individualized characteristics and styles.
In short, most current stochastic and goal-driven approaches might ignore agents' multi-style characteristics, making them challenging to reflect agents' stylized future choices within limited samplings.

\subsection{Trajectory Categorization Approaches}
Several researchers have tried to model interactions in a refined manner through trajectory classification strategies.
For instance, Deo \ETAL~\cite{deo2018would} classify car motions into lane-passing, overtaking, cutting-in, and drifting-into-ego-lane maneuver classes for freeway traffic.
Moreover, Kothari \ETAL~\cite{kothari2020human} propose an exact trajectory categorization method and divide pedestrian trajectories into static, linear, leader-follower, collision-avoidance, group, other interactions, and non-interacting categories when modeling social interaction.
Kothari \ETAL~\cite{kothari2020human} also annotate all the training data in the entire dataset into the above categories, thus achieving the purpose of learning interaction relationships and trajectory plannings categorically.
Kotoari \ETAL~\cite{kothari2021interpretable} also divide agents' interactive behaviors into avoid-occupancy, keep-direction, leader-follower, and collision-avoidance, therefore modeling interactions according to these ``behavior labels''.

It should be noted that some methods have used classification results to describe interactions categorically, but not agents' behaviors.
In addition, these manual classification strategies may increase the workload of model training, and it may be challenging to adapt to different prediction scenarios.
\\

Inspired by these approaches, we wish to employ trajectory classification to describe agents' multi-style future decision-making process through an adaptive classification strategy.
In detail, we first classify agents into several hidden behavior categories according to their destination planning and interactive context.
Then, we train the corresponding feature extraction network for each category in parallel, thus acquiring multiple latent feature spaces with distributional differences.
Finally, the network could simultaneously make multiple future predictions with sufficient categorical style differences by randomly sampling features in \emph{multiple} latent distributions.
Notably, we do not expect to use the quantitative evaluation of the ``better'' or ``worse'' for each style's prediction.
Instead, we want to learn multiple differentiated potential latent distributions through the categorical way, thus reflecting agents' multi-style future choices that have ``equal'' status adaptively.

\section{\MODEL}

\begin{table}[tbp]
    \renewcommand{\arraystretch}{1.3}
    \caption{
        Layers used in the proposed \MODEL.
    }
    \label{tab_method}
    \centering

    \begin{tabular}{c|l}
        \hline
        Layers & Structure (Symbols with * are temporary variables) \\
        
        \hline
        MLP$_e$ & $\bm{X} \to \mbox{fc}(d/2) \to \tanh() \to \bm{f}_t$ \\

        \hline
        \multirow{2}{*}{MLP$_c$} & \makecell[l]{
            $\bm{C} \to$ AvgPooling$(5 \times 5)$ $\to$ Flatten() 
            $\to \mbox{fc}(dt_h) \to $ 
        }\\
        & \makecell[l]{ $\tanh()$ $\to$ Reshape($t_h \times d$) $\to \bm{f}_c$
        }\\

        \hline
        \multirow{2}{*}{Tran$_b$} & \makecell[l]{
        $\bm{f} \to$ TransformerEncoder ($d$) $\to \bm{f}1^*$; 
        $\bm{f}1^*, \bm{X}$ $\to$ 
        }\\
        & \makecell[l]{ TransformerDecoder ($d$)
        $ \to \mbox{fc}(d) \to \bm{h}_\alpha $
        }\\

        \hline
        \multirow{2}{*}{$G$} & \makecell[l]{
            $\bm{h}_\alpha \to K_c \times \mbox{Conv}(1 \times t_h) \to \mbox{fc}(d) \to \bm{F}$; 
        } \\
        & \makecell[l]{$\bm{F} \to \tanh() \to \mbox{fc}(2) \to \bm{D}_p$}\\

        \hline
        \multirow{2}{*}{Tran$_i$} & \makecell[l]{
            $\bm{f}_k \to $ TransformerEncoder($d$) $\to \bm{f}_2$; 
        }\\
        & \makecell[l]{$\bm{f}_2, \hat{\bm{Y}}_l, \bm{X} \to$ TransformerDecoder($d$)
        $\to \bm{h}_\beta$}\\

        \hline
        MLP$_d$ & \makecell[l]{
            $\bm{h}_\beta \to \mbox{fc}(2) \to \hat{\bm{Y}}_D$
        }\\

        \hline
        \multirow{2}{*}{MLP$_g$} & \makecell[l]{
            Concat$(\bm{h}_\beta, \bm{z}) \to \mbox{fc}(d) \to \mbox{ReLU}() \to \bm{g}1^*$;
        }\\
        & \makecell[l]{Concat$(\bm{h}_\beta, \bm{z}) + \bm{g}1^* \to \mbox{fc}(2) \to \hat{\bm{Y}}_g$}\\
        
        \hline
    \end{tabular}
    
\end{table}

\begin{figure*}[t]
    \centering
    \includegraphics[width=1.0\textwidth]{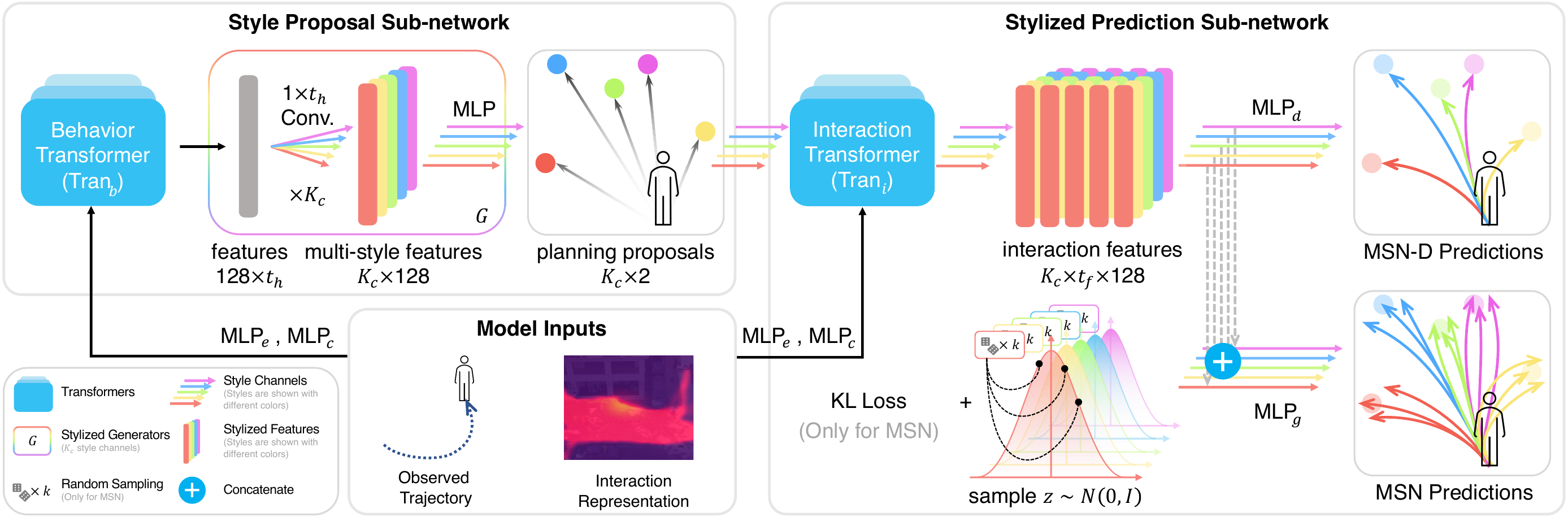}
    \caption{
        \MODEL~has two sub-networks, the style proposal and the stylized prediction sub-network.
        It aims to forecast multi-style future trajectories via $K_c$ style channels (shown with arrows and features in different colors).
        Each stylized trajectory is obtained through the inference of the corresponding style channel.
    }
    \label{fig_overview}
\end{figure*}

This paper focuses on forecasting agents' trajectories in a novel multi-style way.
The proposed model contains two main prediction stages.
We utilize two Transformer-based sub-networks, the \emph{style proposal} and the \emph{stylized prediction} sub-networks, to achieve the multi-style prediction goal with the optimization of the novel stylized loss.
\FIG{fig_overview} shows the overall architecture of \MODEL, and \FIG{fig_stylized_loss} shows how the stylized loss works.
Detailed structures and layers are listed in \TABLE{tab_method}.

\subsection{Problem Formulations}
Given a video clip $\{\bm{I}\}$ that contains $N_a$ agents along with their recorded trajectories during the past $t_h$ observed frames, trajectory prediction takes into account trajectories and interactions as well as scene constraints to predict their future $t_f$ frames' possible positions.
We denote the location of agent-$i$ at frame $t$ as $\bm{p}^i_t = ({p_{x}}^i_t, {p_{y}}^i_t) \in \mathbb{R}^2$.
To distinguish coordinates at observation and prediction periods, we denote $\bm{x}^i_t = \bm{p}^i_t$ when $1 \leq t \leq t_h$, and $\bm{y}^i_t = \bm{p}^i_t$ when $t_h + 1 \leq t \leq t_h + t_f$.
Agent-$i$'s observed trajectory can be written as $\bm{X}^i = \left\{\bm{x}^i_t\right\}_{t=1}^{t_h}$, and the future groundtruth trajectory as $\bm{Y}^i = \left\{\bm{y}^i_t\right\}_{t=t_h+1}^{t_h + t_f}$.
Trajectory prediction focused in this study is to predict agents' possible future positions $\mathcal{Y} = \{\hat{\bm{Y}}^i\}_{i=1}^{N_a}$ according to their observed trajectories $\mathcal{X} = \{\bm{X}^i\}_{i=1}^{N_a}$ and the scene context got from RGB image $\bm{I}$.
Note that the subscript $i$ that denotes the agent's index will be omitted for a simple expression.

\subsection{Style Proposal Sub-network}

The style proposal sub-network aims to model agents' various behavior styles into several \emph{Hidden Behavior Categories}, and then makes planning proposals by assuming agents may behave like each potential behavior style.

\noindent\textbf{a. Behavior Features.}
Agents always consider their current status and their surroundings to plan their activities.
Following \cite{xia2022cscnet}, we use agents' observed trajectories and their transferred images\footnote{
    Xia \ETAL~\cite{xia2022cscnet} propose transferred image $\bm{C}$ to model agents' interactive context, which is computed through the scene RGB images $\bm{I}$ via a CNN.
} together to represent their interactive status.
We employ an MLP (Multi-layer Perceptron) named MLP$_{e}$ (see details in \TABLE{tab_method}) to embed the agent's observed trajectory $\bm{X}$ into the $d/2$ dimensional feature, and take another $\mbox{MLP}_{c}$ to model interactions into the context feature $\bm{f}_c$.
We obtain agent representations by concatenating these two features.
Formally,
\begin{equation}
    \begin{aligned}
        \label{eq_alpha_concat}
        \bm{C} &= \mbox{CNN}(\mathcal{X}, \bm{I}), \\
        \bm{f}_t &= \mbox{MLP}_{e}(\bm{X}) \in \mathbb{R}^{t_h \times d/2}, \\
        \bm{f}_c &= \mbox{MLP}_{c}(\bm{C}) \in \mathbb{R}^{t_h \times d/2}, \\
        \bm{f} &= \mbox{Concat}(\bm{f}_t, \bm{f}_c) \in \mathbb{R}^{t_h \times d}.
    \end{aligned}
\end{equation}

We employ a Transformer-based behavior encoder (Tran$_b$) to infer agents' behavior features $\bm{h}_\alpha$ from their representations $\bm{f}$ and the observed trajectories $\bm{X}$.
Agents' representations $\bm{f}$ will be fed into the Transformer Encoder to calculate the self-attention feature.
After that, the trajectories $\bm{X}$ will query the self-attention feature in the Transformer Decoder to get the final behavior features $\bm{h}_\alpha$.
Formally,
\begin{equation}
    \bm{h}_\alpha = \mbox{Tran}_b(\bm{f}, \bm{X}) \in \mathbb{R}^{t_h \times d}.
\end{equation}

\noindent\textbf{b. Behavior Classification.}
We try to model agents' multi-style characteristics by utilizing an adaptive way to classify all prediction samples into several \emph{Hidden Behavior Categories}.
As discussed above, the mentioned ``style'' indicates agents' continuously changing preferences.
Inspired by recent \emph{goal-driven} trajectory prediction approaches \cite{mangalam2020not}, we take agents' trajectory \emph{end-points} as one of the manifestations to characterize agents' behavior styles.
In detail, we regard that agents with similar enough observations, interactive context, and similar end-point plans belong to the same behavior category.

Accordingly, we introduce a simple similarity measurement method to judge whether different prediction samples (with the same observations and interactive context) belong to the same style category.
We use the nearest neighbor classification method.
Specifically, we distinguish their categories by finding the 2D end-point $\bm{D}_s$ from a set of other end-point proposals $\{\bm{D}_i\}_i = \{(d^i_x, d^i_y)\}_i$, which could make the Euclidean distance between the target 2D end-point $\bm{d} = (d_x, d_y)$ and the selected $\bm{D}_s$ reach the minimum value.

In addition, we assume that these end-point proposals should already be labeled.
Specifically, we use a 2-tuple $(\bm{D}_s, c_s)$ to denote the label $c_s$ of the corresponding end-point $\bm{D}_s$.
Thus, the target trajectory with the end-point $\bm{d}$ and another trajectory with an end-point proposal $\bm{D}_s$ will be classified with the same label $c_s$.
Given the set of 2-tuples $\mathcal{D} = \left\{(\bm{D}_i, c_i)\right\}_i$ and the trajectory with a end-point $\bm{d}$ to be classified, we have
\begin{equation}
    \label{eq_alpha_loss1}
    \begin{aligned}
        \mbox{Category}(\bm{d}|\mathcal{D}) &= c_s, ~
        \mbox{where}~s = \underset{i}{\arg\min} \Vert \bm{D}_i - \bm{d} \Vert.
    \end{aligned}
\end{equation}

It is a relative classification strategy that can be adapted to various prediction scenarios and agents with different behavior preferences.
In addition, the minimum operation also saves complicated work like setting and verifying thresholds.

\begin{figure*}[tbp]
    \centering
    \includegraphics[width=1.0\textwidth]{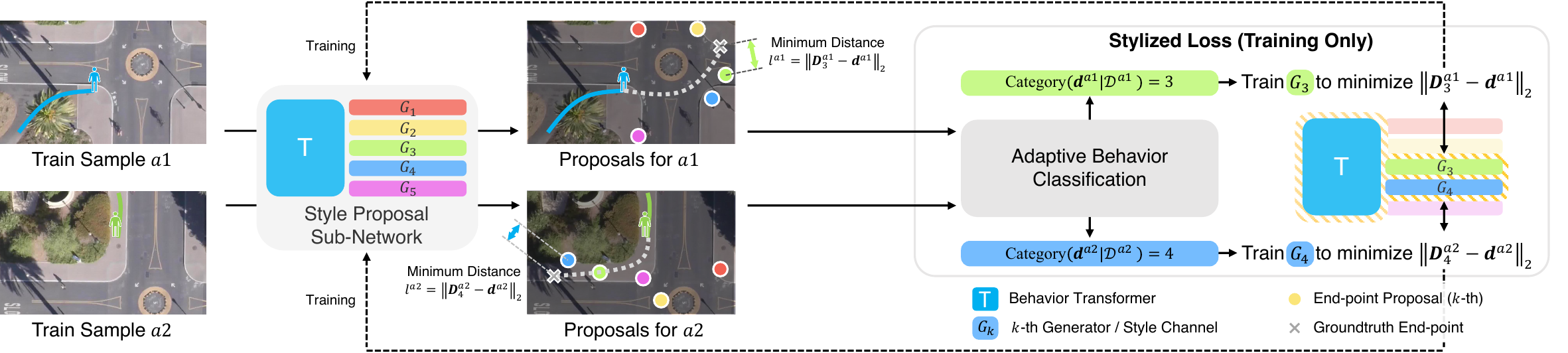}
    \caption{
        Illustration of the behavior classification and the stylized loss.
        When training the network, each prediction sample will be classified by considering the Euclidean distance between end-point proposals $\{\bm{D}_1, \bm{D}_2, ..., \bm{D}_{K_c}\}$ and the groundtruth end-point $\bm{d}$.
        The stylized loss is used to minimize the above minimum distance by tuning trainable parameters in both the behavior Transformer and the corresponding style generator $G_k$.
    }
    \label{fig_stylized_loss}
    
\end{figure*}

~\\\noindent\textbf{c. Style Proposals.}
In \EQUA{eq_alpha_loss1}, we mentioned that to classify a end-point $\bm{d}$, we need a set of end-point proposals $\{\bm{D}_i\}_i$ and their corresponding labels $\{c_i\}_i$.
In the scenario of trajectory prediction, labels $\{c_i\}_i$ may be difficult to obtain directly.
Therefore, we accomplish multi-style prediction by generating several end-point proposals and assigning ``virtual labels'' to them.
We first discuss how to obtain multi-style end-point proposals.
We employ a set of generators $G = \{G_1, G_2, ..., G_{K_c}\}$ to generate end-point proposals with different styles correspondingly.
Each generator could focus on different patterns that lead to the specific behavior style rather than the ``common'' behavior representations.
In detail, we employ $K_c$ convolution kernels $\mathcal{K} = \{\bm{ker}_i\}_{i=1}^{K_c}$ (each kernel has the shape $1 \times t_h$) as well as their corresponding MLPs to encode the multi-style features into multiple latent spaces from agents' behavior features.
Convolution operations are applied on behavior features $\bm{h}_{\alpha}$ to obtain features $\bm{F}$ from $K_c$ channels with strong discriminative categorical styles.
Formally,
\begin{equation}
    \label{eq_alpha_feature}
    \begin{aligned}
        \bm{F} = \mbox{MLP}\left(\mbox{Conv}(\bm{h}_{\alpha}^T, \mathcal{K})\right) \in \mathbb{R}^{K_c \times d}.
    \end{aligned}
\end{equation}
Another decoder MLP is employed to infer the multi-style planning (trajectory end-point) proposals $\bm{D}_p$.
Formally,
\begin{equation}
    \label{eq_D}
    \bm{D}_p = \mbox{MLP}(\bm{F}) \in \mathbb{R}^{K_c \times 2}.
\end{equation}

We combine the above convolution layers and MLPs as the style generator $G$.
Due to the convolution operation with different channels, $G$ contains $K_c$ style channels, and each channel can be regarded as a stylized sub-generator $G_k$.
Each sub-generator will be randomly initialized to ensure they can output different proposals before the training process.
Accordingly, \EQUA{eq_alpha_feature} and \ref{eq_D} can be rewritten together as:
\begin{equation}
    \left\{\begin{aligned}
        \bm{D}_k &= G_k(\bm{h}_\alpha, \bm{ker}_k), \\
        \bm{D}_p &= (\bm{D}_1, ..., \bm{D}_k, ..., \bm{D}_{K_c})^T = G(\bm{h}_\alpha, \mathcal{K}).
    \end{aligned}\right.
\end{equation}
Here, $\bm{D}_k \in \mathbb{R}^2~(k=1, 2, ..., K_c)$ represents the end-point proposal output from the $k$-th style channel.

~\\\noindent\textbf{d. Stylized Loss.}
We now discuss labeling these end-point proposals and training the corresponding sub-generators.
As the groundtruth end-point $\bm{d}$ is known when training, we assign the ``virtual labels'' to the corresponding set of proposals $\{\bm{D}_i\}_{i=1}^{K_c}$ to start the model training for each style channel.
Simply, for the 2-tuple $(\bm{D}_k, c_k)$, we have ``virtual labels'':
\begin{equation}
    \label{eq_alpha_class}
    \begin{aligned}
        c_k = k, ~~k = 1, 2, ..., K_c.
    \end{aligned}
\end{equation}

Given the joint distribution (marked with $P_{\bm{D}_k}$) of the end-point proposals $\bm{D}_k$ and groundtruths $\bm{d}$ that are classified into the $k$-th $(k=1, 2, ..., K_c)$ style category, the corresponding $k$-th generator $G_k$ will be trained through the \emph{stylized loss function} $\mathcal{L}_{sty}(k)$.
It aims to minimize the Euclidean distance between each proposal $\bm{D}_k$ and groundtruth end-point $\bm{d}$ with the label $k$.
Given 2-tuples $\mathcal{D} = \{(\bm{D}_i, c_i)\}_{i=1}^{K_c}$, we have
\begin{align}
    P_{\bm{D}_k} &= P \left(\bm{D}_k, \bm{d}~|~\mbox{Category}(\bm{d} | \{(\bm{D}_i, c_i)\}_i) = k\right), \\
    \mathcal{L}_{sty} (k) &= \mathbb{E}_{(\bm{D}_k, \bm{d}) \sim {P_{\bm{D}_k}}} \Vert \bm{D}_k - \bm{d} \Vert.
    \label{eq_alpha_loss2}
\end{align}

As shown in \FIG{fig_stylized_loss}, each style channel in the style proposal sub-network will be tuned with the corresponding stylized loss when training.
The overall optimization goal is the superposition of all $K_c$ different $L_{sty}(k)$.
Before training, each stylized generator $G_k$ (style channel) will be randomly initialized to ensure the whole network can forecast $K_c$ ($K_c = 5$ in the above schematic figure) different end-point proposals to the target agent.
Then, they learn how to adaptively classify samples into $K_c$ categories, learn how each category's behavior features distribute, and further learn different behavior patterns that different style channels should focus on.
Finally, the sub-network could output multiple proposals with $K_c$ different styles in parallel from the corresponding channels.

\subsection{Stylized Prediction Sub-network}

The stylized prediction sub-network aims to forecast complete trajectories under the given multi-style end-point proposals and the interaction details.
It also considers agents' uncertain future choices (the \MODELG).

\noindent\textbf{a. Interaction Features.}
Similar to \EQUA{eq_alpha_concat}, we take agents' observed trajectories $\bm{X}$ and the transferred images $\bm{C}$ to represent their status and interactions.
We expand the number of time steps from $t_h$ into $t_h + 1$ to fit the stylized planning proposals.
Meanwhile, the MLP$_c$ has become the MLP$_c^*$, where its $t_h$ has been replaced by $t_h + 1$.
For $k$-th proposal $\bm{D}_k$, we have agents' interaction representation $\bm{f}_k$:
\begin{equation}
    \begin{aligned}
        {\bm{f}_t}_k &= \mbox{MLP}_{e}(\mbox{Concat}(\bm{X}, \bm{D}_k)), \\
        {\bm{f}_c}_k &= \mbox{MLP}_{c}^*(\bm{C}), \\
        \bm{f}_k &= \mbox{Concat}({\bm{f}_t}_k, {\bm{f}_c}_k) \in \mathbb{R}^{(t_h + 1) \times d}.
    \end{aligned}
\end{equation}

Similar to the above behavior features, we take the Transformer as the backbone of the interaction encoder (Tran$_i$) to obtain interaction features.
When an agent has specific future planning and no other restrictions, the realistic plan is to move toward its destination straightly.
Other factors affecting its movement can be considered an embodiment of interactive behaviors.
We use the linear spatio-temporal interpolation sequence between agents' current positions (the last observed point) and the given end-point proposal as one of the inputs to model these interactive behaviors.
Formally, the linear sequence $\hat{\bm{Y}}^{l}_{k}$ and the interaction feature ${h_\beta}_k$ are computed through the $k$-th style channel as:
\begin{align}
    \hat{\bm{Y}}^{l}_{k} &= {\left\{\bm{x}_{t_h} + \frac{t}{t_f} (\bm{D}_k - \bm{x}_{t_h})\right\}_{t=1}^{t_f}}^T \in \mathbb{R}^{t_f \times 2}, \\
    {\bm{h}_\beta}_k &= \mbox{Tran}_i \left(\bm{f}_k, \mbox{Concat}\left(\bm{X}, \hat{\bm{Y}}^{l}_{k}\right)\right) \in \mathbb{R}^{t_f \times d}.
\end{align}
The Tran$_i$ eventually outputs the multi-style interaction representations $\bm{h}_{\beta}$ by implementing on all $K_c$ style channels\footnote{All style channels' weights are shared in the stylized prediction stage to save computation resources in our experiments.}:
\begin{equation}
    \bm{h}_{\beta} = \left({\bm{h}_\beta}_1, {\bm{h}_\beta}_2, ..., {\bm{h}_\beta}_{K_c}\right)^T \in \mathbb{R}^{K_c \times t_f \times d}.
\end{equation}

~\\\noindent\textbf{b. Multi-Style Prediction.}
\MODEL~finally decodes features from multiple latent representations $\bm{h}_{\beta_k}$ to provide multiple predicted trajectories with the help of all style channels.
To handle different prediction scenarios, we provide two kinds of \MODEL, the deterministic \MODELD~and the stochastic \MODELG.

(i) \emph{\MODELD}:
We employ an MLP (MLP$_d$) to obtain the multi-style deterministic 2D-coordinate prediction sequences $\hat{\bm{Y}}_D$ from the interaction features $\bm{h}_\beta$.
Formally,
\begin{equation}
    \hat{\bm{Y}}_D = \mbox{MLP}_d(\bm{h}_\beta) \in \mathbb{R}^{K_c \times t_f \times 2}.
\end{equation}
Trajectories with different styles are implemented via different channels (weights are shared).
Then, \MODELD~could forecast $N = K_c$ trajectories with $K_c$ styles from one implementation without repeated sampling operations.

(ii) \emph{\MODELG}:
Deterministic models could hardly reflect agents' uncertainty or randomness when making decisions.
We expand the original MLP$_{d}$ into the MLP$_g$ to generate stochastic predictions \emph{within} each latent space to bring predictions randomness factors.
The \MODELG~will be trained with the addition \textbf{KL loss} (Kullback-Leibler Divergence), thus allowing it to make random predictions via the noise vector.
Given a random sampled vector $\bm{z} \sim N(\bm{0}, \bm{I})$, we have:
\begin{equation}
    \hat{\bm{Y}}_g = \mbox{MLP}_g(\bm{h}_\beta, \bm{z}) \in \mathbb{R}^{K_c \times t_f \times 2}.
\end{equation}
After repeating for $k$ times, we have the stochastic predictions $\hat{\bm{Y}}_G = \left(\hat{\bm{Y}}_{g1}, \hat{\bm{Y}}_{g2}, ..., \hat{\bm{Y}}_{gk}\right)^T \in \mathbb{R}^{k \times K_c \times t_f \times 2}$.
\MODELG~could output $N = kK_c$ trajectories after $k$ times of random sampling.

\subsection{Loss Functions}

We train the \MODEL~end-to-end with the loss function:
\begin{equation}
    \label{eq_loss}
    \mathcal{L}_{\mbox{\MODEL}} = \underbrace{\mathcal{L}_{\mbox{ms}}}_{\mbox{stage 1}} + \underbrace{\mathcal{L}_{\mbox{ad}} + \mathcal{L}_{kl}}_{\mbox{stage 2}}.
\end{equation}

\textbf{Multi-Style Loss}
$\mathcal{L}_{\mbox{ms}}$ is used to train the stage 1 sub-network, making each of the style channels available to output proposals with different styles.
It combines the stylized losses $\mathcal{L}_{sty}(k)$ of all style channels ($k = 1, 2, ..., K_c$).
See detailed explanations in \EQUA{eq_alpha_loss1} and \EQUA{eq_alpha_loss2}.
Formally,
\begin{equation}
    \mathcal{L}_{\mbox{ms}} = \mathbb{E}_{k \sim P(k)} \mathcal{L}_{sty} (k).
\end{equation} 

\textbf{Average Displacement Loss}
$\mathcal{L}_{\mbox{ad}}$ is the average point-wise displacement for each predicted point.
Formally,
\begin{equation}
    \mathcal{L}_{\mbox{ad}} = \frac{1}{N_at_f} \sum_i \sum_t \Vert \bm{y}^i_t - {\hat{\bm{y}}}^i_t \Vert.
\end{equation}

\textbf{KL Loss}
$\mathcal{L}_{kl}$ is the KL divergence between the distribution of agent features $\bm{h}_\beta$ (denoted by $P_\beta$) and the normalized Gaussian distribution $N(\bm{0}, \bm{I})$.
Formally,
\begin{equation}
    \mathcal{L}_{kl} = D_{kl}(P_\beta||N(\bm{0}, \bm{I})), 
\end{equation}
where $D_{kl}(A||B)$ denotes the KL Divergence between distribution $A$ and $B$.
It is used to train the MLP$_g$ in \MODELG~only.

\section{Experiments}

\subsection{Datasets}

We evaluate \MODEL~on three public available trajectory datasets, including ETH\cite{youWillNeverWalkAlone}, UCY\cite{2007Crowds}, and Stanford Drone Dataset (SDD) \cite{learningSocialEtiquette}.
They contain trajectories with rich social and scene interactions in various heterogeneous scenarios.

\textbf{\ETHUCY~Benchmark.}
The \ETHUCY~benchmark has been widely used to evaluate prediction performance.
It contains five video clips in several different scenarios: eth and hotel from ETH, and univ, zara1, zara2 from UCY.
Its annotations are pedestrians' real-world coordinates (in meters) with a specific sampling interval.
We follow the widely used ``leave-one-out'' strategy in previous studies \cite{socialLSTM,socialGAN} when training and evaluating the proposed methods on \ETHUCY.

\textbf{Stanford Drone Dataset.}
The Stanford Drone Dataset (SDD) \cite{learningSocialEtiquette} is a popular dataset for object detection, tracking, trajectory prediction, and many other computer vision tasks.
Many recent \SOTA~trajectory prediction methods \cite{mangalam2020not,liang2020simaug} start to evaluate their ideas on it.
It contains 60 bird-view videos captured by drones over Stanford University.
Positions of more than 11,000 different agents with various physical types, \EG, pedestrians, bicycles, and cars, are given through bounding boxes in pixels.
It has over 185,000 interactions between agents and over 40,000 interactions among agents and scene objects \cite{learningSocialEtiquette}.
Compared with \ETHUCY, it has a richer performance in interaction complexity and differences in scene appearance or structure.
Dataset split used to train \MODEL~on SDD contains 36 training videos, 12 validation videos, and 12 test videos\cite{liang2020simaug,Liang_2020_CVPR}.

\subsection{Implementation Details}

We train and evaluate our model by predicting agents' future coordinates in $t_f = 12$ frames according to their observed $t_h = 8$ frames' coordinates along with the video context.
The frame rate is set to 2.5 fps, \IE, the sample interval is set to $\Delta t = 0.4$s.
We train the entire network with the Adam optimizer with a learning rate $lr = 0.0003$ on one NVIDIA Tesla P4 graphic processor.
In detail, we train \MODEL s~800 epochs on \ETHUCY, and 150 epochs on SDD.
When making training samples, we use a rectangular sliding window with the bandwidth = 20 frames and stride = 1 frame to process original dataset files to obtain training samples \cite{socialGAN}.
The feature dimension is set to $d = 128$.
Detailed layers' connections and activations used are listed in \TABLE{tab_method}.

\subsection{Metrics, Sampling, and Baselines}

\textbf{Metrics.}
We use the Average Displacement Error (ADE) and the Final Displacement Error (FDE) to measure network performance with the \emph{best-of-20} validation \cite{socialGAN,9686621}, which takes the minimum ADE and FDE among 20 predictions to measure prediction accuracy.
ADE is the average point-wise $\ell_2$ error between prediction and groundtruth, and FDE is the $\ell_2$ error of the last prediction point.
For one prediction $\hat{\bm{Y}} = \{\hat{\bm{y}}_t\}_{t = t_h + 1}^{t_h + t_f}$ and its groundtruth $\bm{Y}$, we have:
\begin{equation}
    \begin{aligned}
        \mbox{ADE}(\bm{Y}, \hat{\bm{Y}}) &= \frac{1}{t_f} \sum_{t=t_h+1}^{t_h+t_f} \left\Vert \bm{y}_t - \hat{\bm{y}}_t \right\Vert, \\
        \mbox{FDE}(\bm{Y}, \hat{\bm{Y}}) &= \left\Vert \bm{y}_{t_h + t_f} - \hat{\bm{y}}_{t_h + t_f} \right\Vert.
    \end{aligned}
\end{equation}

\textbf{Sampling.}
Since different models may use different trajectory sampling and generation strategies, we define the \emph{sampling} in this paper as ``the number of samplings (noise or features) \TO the number of generated trajectories''.
Under this definition, we have the sampling ``1 \TO $K_c$'' for \MODELD, and ``$k$ \TO $kK_c$'' for \MODELG~that implements $k$ times.
To compare \MODEL~with others, the $K_c$ is set to $20$, and the $k$ is set to $1$ to align with the \emph{best-of-20} validation.

\textbf{Baselines.}
We choose several \SOTA~methods as our baselines, including 
\SOCIALGAN \cite{socialGAN},
\SRLSTM \cite{srLSTM},
\SOPHIE \cite{sophie},
\BIGAT \cite{bigat},
\PEEKING \cite{peekingIntoTheFuture},
\STGAT \cite{stgat},
\STGCNN \cite{stgcnn},
\GARDEN \cite{Liang_2020_CVPR},
\TRAJECTRONPP \cite{salzmann2020trajectron},
\SIMAUG \cite{liang2020simaug},
\TRANSFORMER \cite{yu2020spatio},
\PECNET \cite{mangalam2020not},
\TPNMS \cite{liang2020temporal},
\SRPAMI \cite{zhang2020social},
\TF \cite{giuliari2020transformer},
\LBEBM \cite{kothari2021interpretable},
\PCCS \cite{Sun_2021_ICCV},
\STC \cite{Li_2021_ICCV},
\YNET \cite{mangalam2020s},
\UNIN \cite{zheng2021unlimited},
\SSALVM \cite{9160982},
\SPECTGNN \cite{cao2021spectral},
\CSC \cite{xia2022cscnet},
\MID \cite{gu2022stochastic},
\MEMO \cite{Xu_2022_CVPR},
\VDRGCN \cite{9686621},
\SSL \cite{tsao2022social},
\NSP \cite{yue2022human},
\SHENET \cite{meng2022forecasting},
\SEEM \cite{wang2022seem}.

\begin{table*}[t]
    \caption{
        Comparisons with the \emph{best-of-20} validation on ETH-UCY.
        Metrics are shown in the format of ``ADE/FDE'' in ``meters''.
        ``D'' denotes deterministic models.
        \DAGGER denotes concurrent work.
        Lower is better.
    }
    \label{tab_ade_all}
    \centering

    \begin{tabular}{c|c|c|ccccc|c}
        \hline
        Models & Source & Sampling & eth & hotel & univ & zara1 & zara2 & Average \\
        
        \hline
        \SOCIALGAN~\cite{socialGAN} & CVPR 2018 & 20 \TO 20 & 0.60/1.19 & 0.52/1.02 & 0.44/0.84 & 0.22/0.43 & 0.29/0.58 & 0.41/0.81 \\
        \SRLSTM~\cite{srLSTM} & CVPR 2019 & 1 \TO 1 (D) & 0.63/1.25 & 0.37/0.74 & 0.51/1.10 & 0.41/0.90 & 0.32/0.70 & 0.45/0.94 \\
        \SOPHIE~\cite{sophie} & CVPR 2019 & 20 \TO 20 & 0.70/0.43 & 0.76/1.67 & 0.54/1.24 & 0.30/0.63 & 0.38/0.78 & 0.54/1.15 \\
        \BIGAT~\cite{bigat} & NeruIPS 2019 & 20 \TO 20 & 0.69/1.29 & 0.49/1.01 & 0.55/1.32 & 0.30/0.62 & 0.36/0.75 & 0.48/1.00 \\
        \PEEKING~\cite{peekingIntoTheFuture} & CVPR 2019 & 20 \TO 20 & 0.73/1.65 & 0.30/0.59 & 0.60/1.27 & 0.38/0.81 & 0.31/0.68 & 0.46/1.00 \\
        \STGAT~\cite{stgat} & ICCV 2019 & 20 \TO 20 & 0.65/1.12 & 0.35/0.66 & 0.52/1.10 & 0.34/0.69 & 0.29/0.60 & 0.43/0.83 \\
        \STGCNN~\cite{stgcnn} & CVPR 2020 & 20 \TO 20 & 0.64/1.11 & 0.49/0.85 & 0.44/0.79 & 0.34/0.53 & 0.30/0.48 & 0.44/0.75 \\
        \TRANSFORMER~\cite{yu2020spatio} & ECCV 2020 & 1 \TO 1 (D) & 0.56/1.11 & 0.26/0.50 & 0.52/1.13 & 0.40/0.89 & 0.31/0.71 & 0.41/0.87\\
        \PECNET~\cite{mangalam2020not} & ECCV 2020 & 20 \TO 20 & 0.54/0.87 & 0.18/0.24 & 0.35/0.60 & 0.22/0.39 & 0.17/0.30 & 0.29/0.48 \\
        \TRAJECTRONPP~\cite{salzmann2020trajectron} & ECCV 2020 & 20 \TO 20 & 0.43/0.86 & \MARK{0.12}/0.19 & \MARK{0.22/0.43} & \MARK{0.17/0.32} & \MARK{0.12/0.25} & \MARK{0.20}/0.39 \\
        \SRPAMI~\cite{zhang2020social} & TPAMI 2020 & 20 \TO 20 & 0.44/0.79 & 0.19/0.31 & 0.50/1.05 & 0.32/0.64 & 0.27/0.54 & 0.34/0.67 \\
        \TPNMS~\cite{liang2020temporal} & AAAI 2021 & 20 \TO 20 & 0.52/0.89 & 0.22/0.39 & 0.55/0.13 & 0.35/0.70 & 0.27/0.56 & 0.38/0.73 \\
        \TF~\cite{giuliari2020transformer} & ICPR 2021 & 20 \TO 20 & 0.61/1.12 & 0.18/0.30 & 0.35/0.65 & 0.22/0.38 & 0.17/0.32 & 0.31/0.55 \\
        \LBEBM~\cite{pang2021trajectory} & CVPR 2021 & 20 \TO 20 & 0.30/0.52 & \MARK{0.13}/0.20 & 0.27/0.52 & 0.20/0.37 & 0.15/0.29 & \MARK{0.21}/0.38 \\
        \PCCS~\cite{Sun_2021_ICCV} & ICCV 2021 & 20 \TO 20 & \MARK{0.28}/0.54 & \MARK{0.11}/0.19 & 0.29/0.60 & 0.21/0.44 & 0.15/0.34 & \MARK{0.21}/0.42 \\
        \STC~\cite{Li_2021_ICCV} & ICCV 2021 & 20 \TO 20 & 0.59/1.12 & 0.20/0.25 & 0.40/0.75 & 0.33/0.55 & 0.27/0.46 & 0.36/0.63 \\
        \YNET~(TTST)~\cite{mangalam2020s} & ICCV~2021 & 10K* \TO 20 & \MARK{0.28/0.33*} & \MARK{0.10/0.14*} & \MARK{0.24/0.41*} & \MARK{0.17/0.27*} & \MARK{0.13/0.22*} & \MARK{0.18/0.27*} \\
        \SSALVM~\cite{9160982} & TCSVT 2021 & 20 \TO 20 & 0.61/1.09 & 0.28/0.51 & 0.30/0.64 & 0.37/0.78 & 0.59/1.24 & 0.43/0.85 \\
        \CSC~\cite{xia2022cscnet} & PR 2022 & 1 \TO 1 (D) & 0.51/1.05 & 0.22/0.42 & 0.36/0.81 & 0.31/0.68 & 0.47/1.02 & 0.37/0.79 \\
        \MID~\cite{gu2022stochastic} & CVPR 2022 & 20 \TO 20 & 0.39/0.66 & \MARK{0.13}/0.22 & \MARK{0.22/0.45} & \MARK{0.17/0.30} & \MARK{0.13/0.27} & \MARK{0.21}/0.38 \\
        \VDRGCN~\cite{9686621} & TITS 2022 & 20 \TO 20 & 0.62/0.81 & 0.27/0.37 & 0.38/0.58 & 0.29/0.42 & 0.21/0.32 & 0.35/0.50 \\
        
        \hline
        \SSL\DAGGER~\cite{tsao2022social} & ECCV 2022 & 1 \TO 1 (D) & 0.69/1.37 & 0.24/0.44 & 0.51/0.93 & 0.42/0.84 & 0.34/0.67 & 0.44/0.85 \\
        \NSP\DAGGER~\cite{yue2022human} & ECCV 2022 & 20 \TO 20 & \MARKR{0.25/0.24\DAGGER} & \MARKR{0.09/0.13\DAGGER} & \MARKR{0.21/0.38\DAGGER} & \MARKR{0.16/0.27\DAGGER} & \MARKR{0.12/0.20\DAGGER} & \MARKR{0.17/0.24\DAGGER} \\
        \SHENET\DAGGER~\cite{meng2022forecasting} & NeurIPS 2022 & 20 \TO 20 & 0.41/0.61 & 0.13/0.20 & 0.25/0.43 & 0.21/0.32 & 0.15/0.26 & 0.23/0.36 \\
        \SEEM\DAGGER~\cite{wang2022seem} & TPAMI 2023 & 20 \TO 20 & 0.62/1.20 & 0.61/1.21 & 0.50/1.04 & 0.31/0.61 & 0.36/0.68 & 0.48/0.95 \\

        \hline
        \MODELD~(Ours)
        & - & 1 \TO 20 (D) & \MARK{0.28}/0.44 & \MARK{0.11/0.17} & 0.28/0.48 & 0.22/0.36 & 0.18/0.29 & \MARK{0.21/0.35} \\
        
        \MODELG~(Ours)
        & - & 1 \TO 20 & \MARK{0.27}/0.41 & \MARK{0.11/0.17} & 0.28/0.48 & 0.22/0.36 & 0.18/0.29 & \MARK{0.21/0.34} \\

        \hline
    \end{tabular}
\end{table*}

\begin{table}[tbp]
    \centering
    \caption{
        Comparisons to baselines with the \emph{best-of-20} on SDD.
        Metrics are shown in the format of ``ADE/FDE'' in pixels.
    }
    \label{tab_ade_sdd}
    \begin{tabular}{c|c|c|c}
        \hline
        Models & Source & Sampling & ADE/FDE \\

        \hline
        \SOCIALGAN & CVPR 2018 & 20 \TO 20 & 27.25/41.44 \\
        \SOPHIE & CVPR 2019 & 20 \TO 20 & 16.27/29.38 \\
        \GARDEN~\cite{Liang_2020_CVPR} & CVPR 2020 & 20 \TO 20 & 14.78/27.09 \\
        \SIMAUG~\cite{liang2020simaug} & ECCV 2020 & 20 \TO 20 & 12.03/23.98 \\
        \TRAJECTRONPP & ECCV 2020 & 20 \TO 20 & 19.30/32.70 \\
        \PECNET & ECCV 2020 & 20 \TO 20 & 9.96/15.88 \\
        \LBEBM & CVPR 2021 & 20 \TO 20 & 8.87/15.61 \\
        \UNIN~\cite{zheng2021unlimited} & CVPR 2021 & 20 \TO 20 & 15.9/26.3 \\
        \SPECTGNN~\cite{cao2021spectral} & ICRA 2021 & 20 \TO 20 & 8.21/12.41 \\
        \YNET~(TTST) & ICCV 2021 & 10K* \TO 20 & 7.85/11.85* \\
        \CSC & PR 2022 & 1 \TO 1 (D) & 14.63/26.91 \\
        \MEMO~\cite{Xu_2022_CVPR}& CVPR 2022 & 20 \TO 20 & 8.56/12.39 \\
        
        \hline
        \NSP\DAGGER & ECCV 2022 & 20 \TO 20 & \MARKR{6.52/10.61\DAGGER} \\
        \SHENET\DAGGER & NeurIPS 2022 & 20 \TO 20 & 9.01/13.24 \\

        \hline
        \MODELD~(Ours) & - & 1 \TO 20 (D) & \MARK{7.69/12.39} \\
        \MODELG~(Ours) & - & 1 \TO 20 & \MARK{7.68/12.16} \\

        \hline
    \end{tabular}
\end{table}

\subsection{Comparison to \SOTABIG~Methods}

For the deterministic \MODELD~and the stochastic \MODELG, we set $K_c = 20$ to meet the \emph{best-of-20} validation, which means that the ``sampling'' of both \MODELD~and \MODELG~are ``1 \TO 20''.
Therefore, the reported results of \MODELD~and \MODELG~are ``similar'' in \TABLE{tab_ade_all} and \TABLE{tab_ade_sdd}.
This section only discusses the performance of \MODELD.
\footnote{
    For detailed results and stochastic performance about the stochastic \MODELG, please refer to the section \ref{sec_stoc} ``Stochastic Performance''.
}
We highlight the works with the best performance with bolded colors, \IE, \MARK{blue} for current works, and \MARKR{purple\DAGGER} for concurrent works.
\footnote{
    If the lowest metric is $v$, we highlight all $v + 0.02$ (meters) on \ETHUCY~and $v + 0.2 $ (pixels) on SDD.
}

\textbf{\ETHUCY.}
Results in \TABLE{tab_ade_all} show that the multi-style deterministic \MODELD~outperforms many current \SOTA~models.
\YNET~reaches the best performance on \ETHUCY~while it needs to sample 10,000 times (Test-Time Sampling Trick, TTST) when generating trajectories.
For a fair comparison, we compare \SOTA~stochastic models \TRAJECTRONPP~\cite{salzmann2020trajectron} and \MID~\cite{gu2022stochastic}.
Compared with \TRAJECTRONPP, \MODELD~significantly improves the ADE and FDE on eth sub-dataset for over 34.8\% and 48.8\%, and on hotel sub-dataset for 8.3\% and 10.5\%.
The average ADE of \TRAJECTRONPP, \MID, and \MODELD~are basically at the same level, but the average FDE of our proposed model gains about 10.2\% on \ETHUCY.
Although \MODELD~does not perform as well as \NSP, it still shows competitiveness compared to other newly published methods.
Note that \MODELD~is \emph{deterministic}, even if it could simultaneously output $N = K_c = 20$ multi-style predictions.
It still lacks uncertain factors when implementing.
Nevertheless, its results on \ETHUCY~show strong competitiveness even with other stochastic models when outputting the same number of trajectories.

\textbf{SDD.}
As shown in \TABLE{tab_ade_sdd}, \MODELD~outperforms current \SOTA~\MID~and \SPECTGNN~\cite{cao2021spectral} on the average ADE on SDD by a significant margin of 7.3\%.
Besides, it has dramatically improved ADE by over 13.3\% and FDE by over 20.6\% compared with \LBEBM.
As a result, \MODELD~performs better than most of the other stochastic models on SDD under the \emph{best-of-20} validation.
It reflects the superiority of \MODELD~in handling complex data since SDD contains more heterogeneous agents and scenarios.

\subsection{Quantitative Analysis}

\begin{table}[btp]
    \caption{
        Ablation Studies on \ETHUCY.
        S1 and S2 represent the style proposal and the stylized prediction.
        Experiment No.1 could not calculate ADE for it only implements S1.
        Results are shown with the \emph{best-of-20} validation.
    }
    \label{tab_ab}
    \centering

    \begin{tabular}{c|ccc}
        \hline
        No. &1&2&3 \\

        \hline
        S1 & \checkmark & \checkmark & \checkmark \\
        S2 & $\times$ & Linear & \checkmark \\

        \hline
        eth & \NA/0.54 & 0.32/0.54 & 0.28/0.44 \\
        hotel & \NA/0.19 & 0.13/0.19 & 0.11/0.17 \\
        univ & \NA/0.52 & 0.30/0.52 & 0.28/0.48 \\
        zara1 & \NA/0.36 & 0.24/0.36 & 0.22/0.36 \\
        zara2 & \NA/0.29 & 0.19/0.29 & 0.18/0.29 \\

        \hline
        Average & \NA/0.38 & 0.23/0.38 & 0.21/0.35 \\
        
        \hline
    \end{tabular}

\end{table}

\begin{table*}[htbp]
    \caption{
        Verifying of $K_c$ in \MODELD~and $k$ in \MODELG.
        Metrics are shown with ``ADE/FDE''.
        $N=kK_c$ denotes the number of output trajectories.
        Results with * are relative values compared to the \MODELD~($K_c = 1$).
        Lower values indicate better performance.
    }
    \label{tab_a_k}
    \centering

    \begin{tabular}{c|cc|c|c|ccccc|cc}
        \hline
        & $k$ & $K_c$ & $N$ & Sampling & eth & hotel & univ & zara1 & zara2 & \ETHUCY & SDD \\ 
        \hline

        \multirow{7}{*}{\rotatebox{90}{\MODELD}}
        & - & 1 & 1 & 1 \TO 1 (D) & 0.65/1.27 & 0.23/0.42 & 0.62/1.23 & 0.46/0.91 & 0.41/0.81 & 0.47/0.93 & 16.76/32.83 \\ 
        & - & 5 & 5 & 1 \TO 5 (D) & 0.43/0.80 & 0.15/0.25 & 0.41/0.74 & 0.31/0.55 & 0.25/0.46 & 0.31/0.56 & 11.30/20.63 \\ 
        & - & 10 & 10 & 1 \TO 10 (D) & 0.37/0.65 & 0.13/0.21 & 0.36/0.65 & 0.28/0.47 & 0.22/0.37 & 0.27/0.47 & 9.52/16.60 \\ 
        & - & 20 & 20 & 1 \TO 20 (D) & 0.28/0.44 & 0.11/0.17 & 0.28/0.48 & 0.22/0.36 & 0.18/0.29 & 0.21/0.35 & 7.69/12.39 \\ 
        & - & 30 & 30 & 1 \TO 30 (D) & 0.27/0.41 & 0.09/0.13 & 0.27/0.43 & 0.18/0.27 & 0.16/0.23 & 0.19/0.28 & 7.12/11.06 \\ 
        & - & 50 & 50 & 1 \TO 50 (D) & 0.24/0.33 & 0.09/0.11 & 0.22/0.32 & 0.17/0.22 & 0.14/0.19 & 0.17/0.23 & 7.13/10.86 \\ 
        & - & 100 & 100 & 1 \TO 100 (D) & 0.27/0.42 & 0.10/0.14 & 0.26/0.46 & 0.20/0.30 & 0.17/0.25 & 0.20/0.27 & 7.58/12.10 \\ 
        
        \hline

        \multirow{5}{*}{\rotatebox{90}{\MODELG}}
        & 1 & 20 & 20 & 1 \TO 20 & 0.27/0.41 & 0.11/0.17 & 0.28/0.49 & 0.23/0.36 & 0.18/0.29 & 0.21/0.34 & 7.68/12.16 \\ 
        & 5 & 20 & 100 & 5 \TO 100 & 0.20/0.27 & 0.08/0.11 & 0.23/0.37 & 0.18/0.24 & 0.14/0.20 & 0.16/0.23 & 5.83/7.93 \\ 
        & 10 & 20 & 200 & 10 \TO 200 & 0.19/0.22 & 0.08/0.09 & 0.22/0.33 & 0.16/0.21 & 0.13/0.17 & 0.15/0.20 & 5.27/6.49 \\ 
        & 20 & 20 & 400 & 20 \TO 400 & 0.18/0.18 & 0.07/0.07 & 0.20/0.29 & 0.15/0.18 & 0.13/0.15 & 0.14/0.17 & 4.86/5.30 \\ 
        & 30 & 20 & 600 & 30 \TO 600 & 0.17/0.16 & 0.07/0.06 & 0.19/0.27 & 0.15/0.16 & 0.12/0.14 & 0.14/0.15 & 4.68/4.71 \\ 
        \hline
    \end{tabular}
\end{table*}

\textbf{a. Multi-Style End-Point Proposals.}
The style proposal sub-network provides $K_c$ end-point proposals to the same agent with multiple planning styles.
Unlike existing goal-driven approaches, the proposed \MODEL~adds an adaptive classification strategy to make it available to make proposals with different planning styles through multiple style channels within one implementation.
As shown in \TABLE{tab_ab}, the average error (FDE) of the proposed end-points even outperforms several current \SOTA~methods (under the \emph{best-of-20} validation).
Please note that we do not calculate ADE since study No. 1 only outputs agents' future end-points rather than the entire trajectories.
Furthermore, study No. 2\footnote{Study No. 2 only uses the simple linear interpolation method to process end-points to obtain the entire predictions.} demonstrates that its performance is even comparable to \TRAJECTRONPP~\cite{salzmann2020trajectron}, \LBEBM~\cite{pang2021trajectory}, \PCCS~\cite{Sun_2021_ICCV}, and \MID~\cite{gu2022stochastic}.
It considers nothing about agents' interactive behaviors and scene constraints.
It shows the excellent performance of the stage 1 style proposal sub-network by adaptively classifying agents into several behavior categories compared to previous goal-driven methods.
Compared to the popular goal-driven \PECNET~\cite{mangalam2020not}, \MODEL~classifies agents into several hidden behavior categories and then makes multiple predictions by assuming agents would behave as each style in parallel, rather than randomly sampling trajectories from the same latent distribution for 20 times to obtain stochastic predictions.
It powerfully demonstrates the overall efficiency of the adaptive classification method and the multi-style prediction strategy.
For further explanations, please refer to the section \ref{sec_sty} ``Prediction Styles''.

\textbf{b. Stylized Prediction.}
As shown in \TABLE{tab_ab}, ADE and FDE of study No. 3 have improved over 7\% compared with study No. 2, which utilizes linear interpolation and considers nothing about any interactions.
It proves the effectiveness of the stylized prediction, which forecasts entire trajectories under the given end-point by considering agents' interactive behaviors.
The most significant difference from No. 2 is that it brings better non-linearity and sociality to the prediction.
In addition, the proposed \MODEL~could bring better multi-style properties by using agents' multiple end-point plannings to characterize their behavioral styles, thus enabling better multi-style generation properties than current goal-driven approaches under the same number of random samplings.

\begin{figure}[tbp]
    \centering
    \includegraphics[scale=0.4]{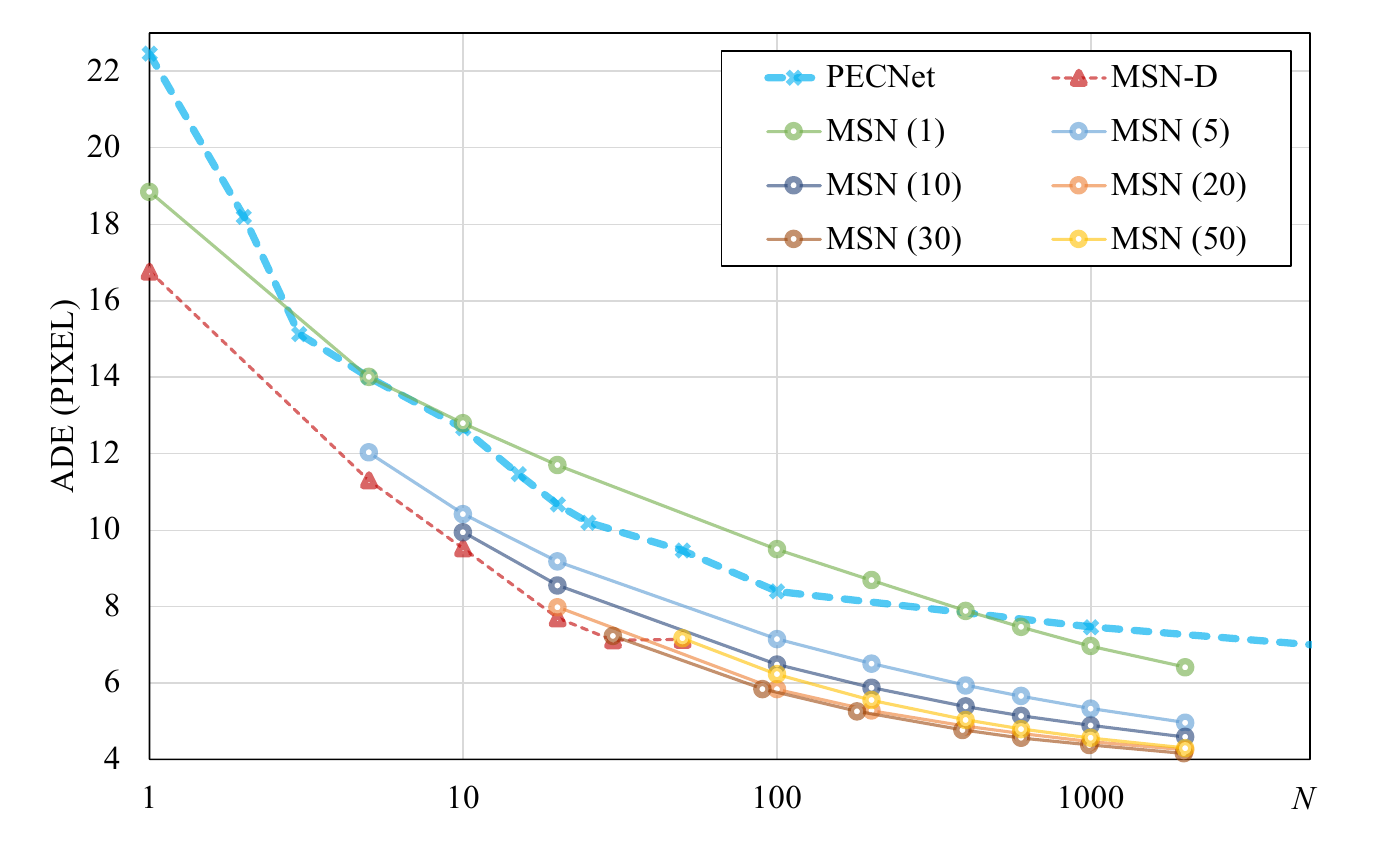}
    \caption{
        ADE of \MODEL~with different $K_c$ and $k$ configures on SDD.
        $N = kK_c$ represents the number of predicted trajectories in once model implementation.
        Number in brackets indicates $K_c$, and we set $K_c = 20$ for the \MODELD.
    }
    \label{fig_ade_k}
\end{figure}

\textbf{c. The Number of Behavior Categories ($K_c$).}
The selection of the hyper-parameter $K_c$ directly affects how \MODEL~``judges'' and classifies agents' potential behavior styles.
Using the adaptive classification strategy means a smaller $K_c$ is not enough to reflect agents' various behavior preferences.
At the same time, a larger $K_c$ might require a large amount of training data to tune each generator to make them available to generate predictions with sufficient style differences.
We run ablation experiments by verifying $K_c$ in $\{1, 5, 10, 20, 30, 50, 100\}$ to explore the stylized performance on \ETHUCY~and SDD.
See details in \TABLE{tab_a_k}.
Results illustrate that \MODEL~performs better with a higher $K_c$ on both \ETHUCY~and SDD when $K_c$ is set to a relatively small value.
Besides, it shows that setting $K_c$ around 30-50 will achieve the best performance on each sub-dataset.
On the contrary, a too-large $K_c$ will cause performance degradation due to the limited training samples and behavior styles on \ETHUCY~and SDD.
It shows that the best $K_c$ setting on \ETHUCY~is about 50, and SDD is about 30.
Nevertheless, we still choose $K_c = 20$ and generate only one trajectory in each behavior category to adapt to the \emph{best-of-20} validation compared to other methods, which means that the performance of the proposed model may not be fully exploited under this evaluation metric.

\textbf{d. Stochastic Performance.}
\label{sec_stoc}
The stochastic \MODELG~samples features from each latent distribution, therefore outputting multiple random predictions.
Random sampling time $k$ matters how the stochastic \MODELG~performs.
We present quantitative comparisons of \MODELG~($K_c = 20$) with several $k$ settings in \TABLE{tab_a_k}.
Besides, to verify the synergy of sampling times $k$ and the number of behavior categories $K_c$ on the stochastic performance of the proposed model, we show the changing of ADE with different $K_c$ and $k$ configures on SDD in \FIG{fig_ade_k}.
We also compare our results with \PECNET~under the same number of output trajectories $N = kK_c$ in \FIG{fig_ade_k}.
It shows that \MODEL~may quickly achieve better performance with a lower $N$.
Compared with the goal-driven \PECNET, \MODELG~improves its ADE for over 41.4\% when generating $N = kK_c = 1000$ trajectories.
It illustrates \MODELG's advantages when generating a large number of predictions.
It should be noted that when generating 1000 outputs, \PECNET~may sample trajectories 1000 times.
Meanwhile, the proposed \MODELG~only takes 50 times of random sampling (when $K_c = 20$), significantly improving the time efficiency, especially when generating a larger number of stochastic predictions.

In addition, it also proves the co-influence of $K_c$ and $k$ when generating the same number of trajectories.
For example, results in \FIG{fig_ade_k} demonstrate that \MODELG~($K_c = 5, k = 4$) performs worse than \MODELG~($K_c = 10, k = 2$) and \MODELG~($K_c = 20, k = 1$), even they generate the same number ($N = kK_c = 20$) of trajectories.
Besides, \MODELD~($K_c = 100$) and \MODELG~($K_c = 20, k = 5$) also show the similar results in \TABLE{tab_a_k} with about 10\% performance difference.
Both \TABLE{tab_a_k} and \FIG{fig_ade_k} indicate that \MODEL~may achieve the best performance when setting $K_c = 30$ on SDD, no matter how many stochastic predictions are generated.
Note that we choose $K_c = 20$ and $k = 1$ as the basic configuration compared to other methods in line with the \emph{best-of-20} validation, which means that the potential of the proposed model could not be fully exploited.
Therefore, the adaptive classification and the multi-style prediction could significantly improve performance.
Moreover, random sampling in multiple latent distributions could help generate more trajectories easily and quickly to suit more application scenarios.

\begin{figure*}[tbp]
\centering
\includegraphics[width=1.0\textwidth]{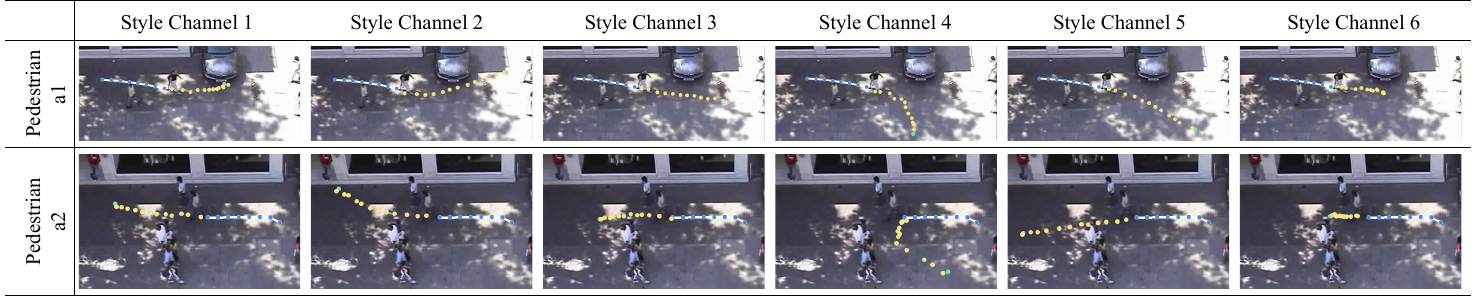}
\caption{
    Visualization of prediction styles.
    We show the first six styles of stylized predictions (yellow dots) for pedestrians a1 and a2 given by \MODELD~($K_c = 20$).
    Their observations are shown with blue dots, and the given end-point proposals with green dots.
}
\label{fig_betamodel}
\end{figure*}

\begin{figure}[tbp]
\centering
\includegraphics[scale=0.25]{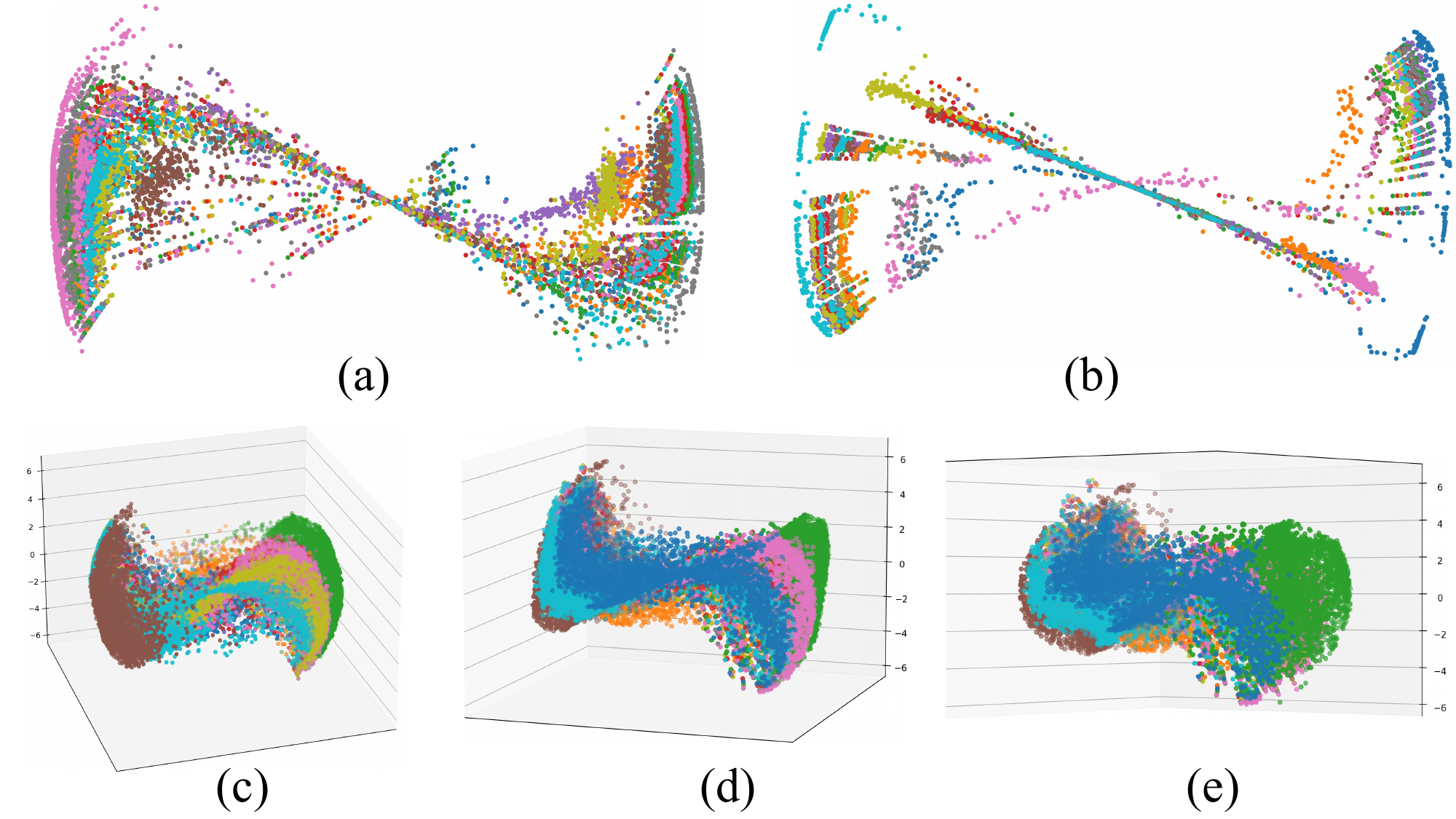}
\caption{
    Stylized features.
    We show the 2D/3D feature distributions of the $d$-dimensional multi-style features via PCA.
    Each dot represents one stylized feature that belongs to a specific hidden behavior category.
    Categories are distinguished by colors.
    We show the 2D feature distributions on eth and zara1 in (a)(b), and the 3D distribution on univ with three views in (c)(d)(e).
}
\label{fig_features}
\end{figure}

\begin{figure*}[t]
\centering
\includegraphics[width=1.0\textwidth]{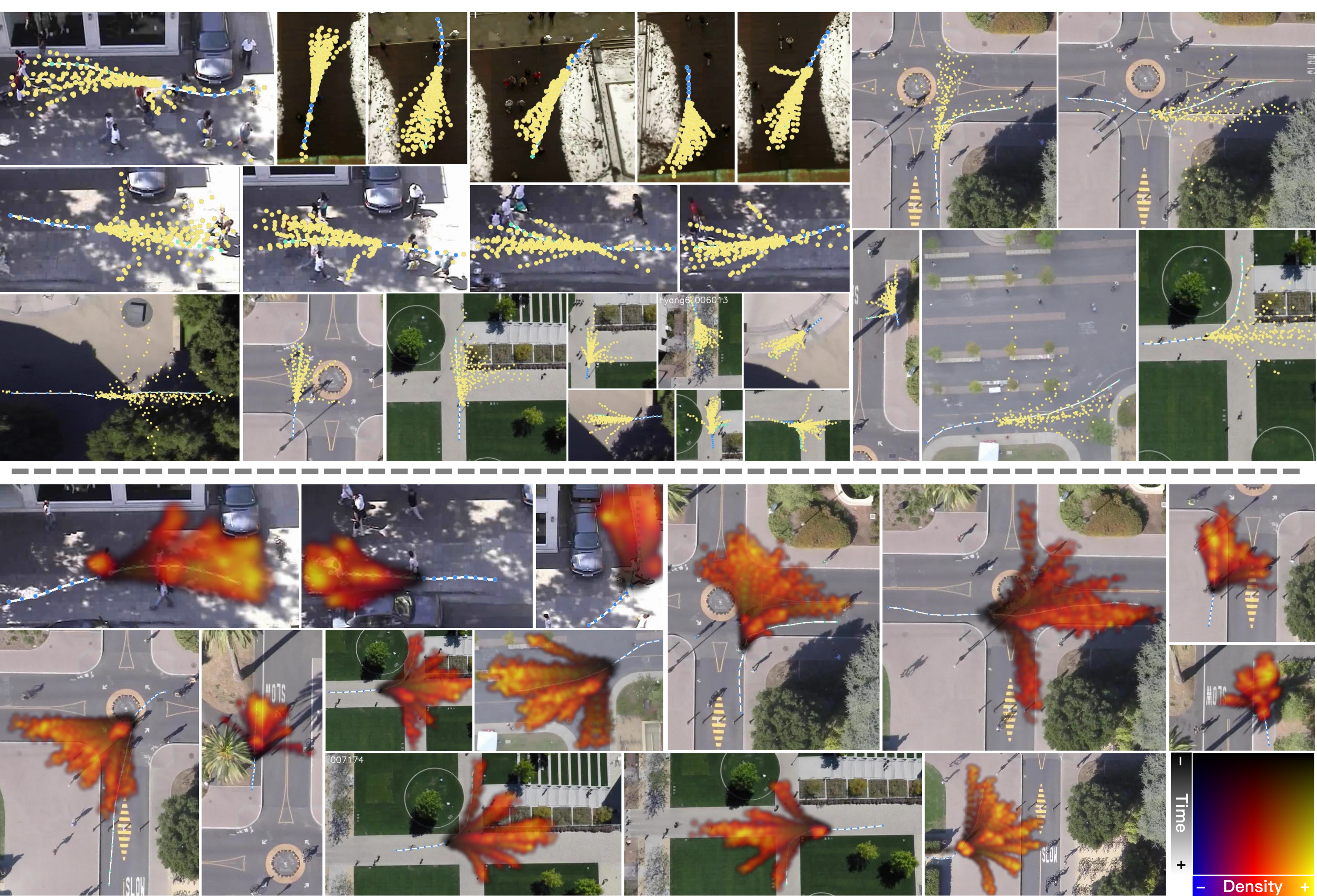}
\caption{
    Results visualization.
    We display visualized results of \MODELD~($K_c = 20$) and \MODELG~(shown with trajectory distributions) among several datasets.
    Bule dotes represent agents' observed trajectories, green dots are their groundtruths, and yellow dots are deterministic predictions given by \MODELD. For heatmaps given by \MODELG, yellow represents higher density and blue represent lower.
    Besides, time steps are distinguished with darker and lighter colors.
}
\label{fig_vis}
\end{figure*}

\subsection{Qualitative Analysis}

\textbf{a. Categorical Feature Visualization.}
We visualize several sampled agents' $d$-dimension multi-style behavior features $\bm{F} \in \mathbb{R}^{K_c \times d}$ to explore their distributions qualitatively.
This distribution of behavior features for all predicted samples in the dataset can be viewed as a superposition of multiple feature distributions in each category.
In other words, the difference in the distribution of behavior features of each category brings the multi-style prediction property.
We use PCA (Principal Components Analysis) to downscale these features to 2D or 3D to study their spatial distribution and distinguish the different categories in different colors.
As shown in \FIG{fig_features}, all the features distribute on a low-dimensional manifold, and each category occupies a specific area.
In other words, the classification strategy adaptively divides the whole feature space into $K_c$ ``sub-spaces'' according to their end-point plannings and interactive context.
Even though we do not know what these ``styles'' real meanings are, \FIG{fig_features} demonstrates that agents' behavior styles could be adaptively classified with such a simple minimum distance measurement.
The stylized generator samples feature from the corresponding latent distribution when forecasting trajectories through the specific style channel.
When generating multiple predictions, we can selectively refer to each stylized distribution prior, rather than by repeated random sampling from the only distribution as in most existing stochastic methods.
It proves the efficiency of adaptive classification and multi-style prediction.

\textbf{b. Prediction Styles.}
\label{sec_sty}
Another significant design is to predict the entire trajectories in multiple style channels under their specific behavior categories.
Specifically, we take agents' end-points and their interactive context to reflect their behavior styles, thereby utilizing $K_c$ style channels to provide the corresponding stylized predictions in parallel.
We show several predictions given by different style channels for pedestrians a1 and a2 in \FIG{fig_betamodel}.
Predictions demonstrate that \MODEL~has robust adaptability to different agents considering their potential agent-agent or agent-scene interactions.
For instance, it makes predictions for a1 and a2 with various interactive behaviors like entering the shop (a2, channel 2), turning around (a1 and a2, channel 4), slowing down (a1 and a2, channel 6), and going around the parking car (a1, channel 2).

Besides, predictions given by the same style channel show similar planning and interaction trends.
For example, style channel 1 makes predictions that both agent a1 and a2 may keep their current motion states and walk forward.
The other style channel 2 considers they may have large-scale interactive behaviors to the scene context, and forecasts agent a1 to get through the parking car and turn around immediately and agent a2 to turn to the shop immediately.
Similarly, channel 4 provides them with trajectories with quite drastic direction changes, and channel 6 thinks they may slow down in the future.
Please note that we did not provide style labels or annotations when training the whole network, and all style channels are trained adaptively with the proposed stylized loss.
It shows the effectiveness of \MODEL~and its strong adaptive ability for generating predictions with multiple behavior styles.

\subsection{Visualization}
We display several prediction results of both \MODELD~and \MODELG~on \ETHUCY~and SDD to show qualitative performance in \FIG{fig_vis}.
Blue dots represent the target agents' observed coordinates, and green dots are their groundtruth future coordinates.
Yellow dots are deterministic predictions given by \MODELD~(outputs totally $N = K_c = 20$ trajectories at once implementation), and heatmaps are the spatial distributions of the predicted trajectories given by \MODELG.
Results demonstrate that the two-stage multi-style \MODEL~has provided multiple styles of predictions, such as pedestrians going ahead, turning around the crossroad, resting, and bicycles turning right or going ahead (but not turning left) at the roundabout.
In the intersection scene shown in the bottom row in \FIG{fig_vis}, it makes a variety of future predictions with sufficient style differences, such as turning right, going forward, turning left, etc.
All the above predictions qualitatively prove the proposed multi-style prediction strategy's effectiveness.

\subsection{Limitations}
Although the proposed \MODEL~achieves better performance, unexpected or failed predictions still exist in some output style channels.
In detail, it may have a few (one or two out of twenty) style channels that predict unexpected stylized predictions in some datasets.
In detail, one or two in all $K_c = 20$ style channels may make not-that-good predictions.
For example, as shown in the first visualized prediction in the top-left of \FIG{fig_vis}, several predictions almost coincide with the observed trajectory (but in the opposite direction).
Although they almost do not affect the quantitative ADE and FDE under the \emph{best-of-20} validation, these predictions violate the physical limitations of agent activities.
We infer that the data of different categories of $K_c$ is unbalanced, leading to incomplete variables tuning in channels corresponding to categories with fewer data.
This phenomenon occurs more in the \ETHUCY~dataset but less in the SDD, which also verifies our point of view.
We will continue studying this problem and solving these failed cases in our future work.

\section{Conclusion}
Forecasting agents' potential future trajectories in crowd spaces is crucial in intelligent transportation systems, but challenging due to agents' uncertain personalized future choices and the changeable scene context.
Many researchers are committed to this task by modeling agents' stochastic interactive behaviors via generative neural networks, or their various intention or destination distributions in the goal-driven approaches.
However, most of them model all kinds of agents' behaviors into a ``single'' latent distribution and randomly sample features to make multiple stochastic predictions.
Thus, giving diversified future predictions with different-enough behavior styles for agents with diverse preferences and personalization is still difficult.
In this paper, we tackle this problem by introducing a novel \emph{multi-style} prediction strategy, therefore trying to make multiple future predictions with sufficient style differences.
We propose \MODEL, a transformer-based multi-style trajectory prediction network to predict agents' trajectories in a multi-style way.
We adaptively divide agents' behaviors into different categories through their interactive context and end-point planning.
Then, we employ and train a series of generators (called the style channels) according to the novel stylized loss constraints.
In this way, different style channels may focus on different patterns of agents' behaviors, thereby sampling features from multiple latent distributions to make differentiated predictions with different planning styles.
Experiments demonstrate that \MODEL~outperforms most of the current \SOTA~prediction models, and can be adapted to various complex prediction scenarios.
The visualized qualitative experimental results also prove the effectiveness of the multi-style prediction strategy.
The proposed models could also provide diversified future choices with sufficient differences that meet the interactive constraints.
In the future, we will continue to explore a suitable way to make differentiated and categorized future predictions for multiple kinds of agents and try to address the failed prediction cases.

\section{Acknowledgment}

This work is supported by National Natural Science Foundation of China (62172177) and Zhuhai Industry-University-Research Cooperation Project (ZH22017001210089PWC).

\bibliographystyle{IEEEtran}
\bibliography{./ref.bib}

\begin{IEEEbiographynophoto}{Conghao Wong}
    received his bachelor's degree from the School of Mechanical Engineering and Electronic Information, China University of Geosciences, Wuhan, China, in 2019, and the master's degree from the School of Electronic Information and Communications, Huazhong University of Science and Technology, Wuhan, China, in 2022. He is currently pursuing the Ph.D. degree. His current research interests include computer vision and pattern recognition.
\end{IEEEbiographynophoto}

\begin{IEEEbiographynophoto}{Beihao Xia}
    received his bachelor's degree from the College of Computer Science and Electronic Engineering, Hunan University, Changsha, China, in 2015, and the master's degree from the School of Electronic Information and Communications, Huazhong University of Science and Technology, Wuhan, China, in 2018. He is currently pursuing the Ph.D. degree. His current research interests include trajectory prediction, behavior analysis and understanding.
\end{IEEEbiographynophoto}

\begin{IEEEbiographynophoto}{Qinmu Peng}
    is currently an Associate Professor at the School of Electronic Information and Communications, Huazhong University of Science and Technology, Wuhan, China. He received his Ph.D. degree from the Department of Computer Science at Hong Kong Baptist University in 2015. His research interests include medical image processing, pattern recognition, machine learning, and computer vision.
\end{IEEEbiographynophoto}

\begin{IEEEbiographynophoto}{Wei Yuan}
    received the BS degree in electronic engineering from Wuhan University, China, in 1999, and the PhD degree in electronic engineering from the University of Science and Technology of China, Hefei, in 2006. He is currently a professor with the School of Electronic Information and Communications, Huazhong University of Science and Technology, China. His current research interests include machine learning and information security.
\end{IEEEbiographynophoto}

\begin{IEEEbiographynophoto}{Xinge You}
    (Senior Member, IEEE) is currently a Professor with the School of Electronic Information and Communications, Huazhong University of Science and Technology, Wuhan.
    He received the B.S. and M.S. degrees in mathematics from Hubei University, Wuhan, China, in 1990 and 2000, respectively, and the Ph.D. degree from the Department of Computer Science, Hong Kong Baptist University, Hong Kong, in 2004.
    His research results have expounded in 180+ publications at prestigious journals and prominent conferences, such as IEEE T-PAMI, T-IP, T-NNLS, T-CYB, T-CSVT, CVPR, ECCV, IJCAI.
    He served/serves as an Associate Editor of the \textit{IEEE Transactions on Cybernetics}, \textit{IEEE Transactions on Systems, Man, Cybernetics: Systems}.
    His current research interests include image processing, wavelet analysis and its applications, pattern recognition, machine earning, and computer vision.
\end{IEEEbiographynophoto}

\vfill

\appendices

\section{Transformer Details}
\label{Appendix_A}
We employ the Transformer \cite{attentionIsAllYouNeed} as the backbone to encode trajectories and the scene context in each of the two sub-networks.
Several trajectory prediction methods (like \cite{yu2020spatio,giuliari2020transformer}) have already tried to employ Transformers as their backbones.
Experimental results have shown excellent improvements.
Transformers can be regarded as a kind of sequence-to-sequence (seq2seq) model \cite{attentionIsAllYouNeed}.
Unlike other recurrent neural networks, the critical idea of Transformers is to use attention layers instead of recurrent cells.
With the multi-head self-attention layers\cite{attentionIsAllYouNeed}, the long-distance items in the sequence could directly affect each other without passing through many recurrent steps or convolutional layers.
In other words, Transformers can better learn the long-distance dependencies in the sequence while reducing the additional overhead caused by the recurrent calculation.
The Transformer we used in the proposed \MODEL~has two main parts, the Transformer Encoder and the Transformer Decoder.
Both these two components are made up of several attention layers.

\subsection{Attention Layers}

Following definitions in \cite{attentionIsAllYouNeed}, each layer's multi-head dot product attention with $H$ heads is calculated as:
\begin{equation}
    \begin{aligned}
        \mbox{Attention}(\bm{q}, \bm{k}, \bm{v}) &= \mbox{softmax}\left(\frac{\bm{q}\bm{k}^T}{\sqrt{d}}\right)\bm{v}, \\
        \mbox{MultiHead}(\bm{q}, \bm{k}, \bm{v}) &= \mbox{fc}\left(\mbox{concat}(\left\{ \mbox{Attention}_i(\bm{q}, \bm{k}, \bm{v}) \right\}_{i=1}^H)\right).
    \end{aligned}
\end{equation}
In the above equation, $\mbox{fc}()$ denotes one fully connected layer that concatenates all heads' outputs.
Query matrix $\bm{q}$, key matrix $\bm{k}$, and value matrix $\bm{v}$, are the three inputs.
Each attention layer also contains an MLP (denoted as MLP$_a$) to extract the attention features further.
It contains two fully connected layers.
ReLU activations are applied in the first layer.
Formally,
\begin{equation}
    \bm{f}_{o} = \mbox{ATT}(\bm{q}, \bm{k}, \bm{v}) = \mbox{MLP}_a(\mbox{MultiHead}(\bm{q}, \bm{k}, \bm{v})),
\end{equation}
where $\bm{q}, \bm{k}, \bm{v}$ are the attention layer's inputs, and $\bm{f}_o$ represents the layer output.

\subsection{Transformer Encoder}

The transformer encoder comprises several encoder layers, and each encoder layer contains an attention layer and an encoder MLP (MLP$_e$).
Residual connections and normalization layers are applied to prevent the network from overfitting.
Let $\bm{h}^{(l+1)}$ denote the output of $l$-th encoder layer, and $\bm{h}^{(0)}$ denote the encoder's initial input.
For $l$-th encoder layer, we have
\begin{equation}
    \label{eq_alpha_encoder}
    \begin{aligned}
        \bm{a}^{(l)} &= \mbox{ATT}(\bm{h}^{(l)}, \bm{h}^{(l)}, \bm{h}^{(l)}) + \bm{h}^{(l)}, \\
        \bm{a}^{(l)}_n &= \mbox{Normalization}(\bm{a}^{(l)}), \\
        \bm{c}^{(l)} &= \mbox{MLP}_e(\bm{a}_n^{(l)}) + \bm{a}_n^{(l)}, \\
        \bm{h}^{(l+1)} &= \mbox{Normalization}(\bm{c}^{(l)}).
    \end{aligned}
\end{equation}

\subsection{Transformer Decoder}

Like the Transformer encoder, the Transformer decoder comprises several decoder layers, and each is stacked with two different attention layers.
The first attention layer focuses on the essential parts in the encoder's outputs $\bm{h}_e$ queried by the decoder's input $\bm{X}$. 
The second is the same self-attention layer as in the encoder.
Similar to \EQUA{eq_alpha_encoder}, we have:
\begin{equation}
    \label{eq_alpha_decoder}
    \begin{aligned}
        \bm{a}^{(l)} &= \mbox{ATT}(\bm{h}^{(l)}, \bm{h}^{(l)}, \bm{h}^{(l)}) + \bm{h}^{(l)}, \\
        \bm{a}^{(l)}_n &= \mbox{Normalization}(\bm{a}^{(l)}), \\
        \bm{a}_2^{(l)} &= \mbox{ATT}(\bm{h}_e, \bm{h}^{(l)}, \bm{h}^{(l)}) + \bm{h}^{(l)}, \\
        \bm{a}_{2n}^{(l)} &= \mbox{Normalization}(\bm{a}_2^{(l)}) \\
        \bm{c}^{(l)} &= \mbox{MLP}_d(\bm{a}_{2n}^{(l)}) + \bm{a}_{2n}^{(l)}, \\
        \bm{h}^{(l+1)} &= \mbox{Normalization}(\bm{c}^{(l)}).
    \end{aligned}
\end{equation}

\subsection{Positional Encoding}

Before inputting agents' representations or trajectories into the Transformer, we add the positional coding to inform each timestep's relative position in the sequence.
The position coding $\bm{f}_e^t$ at step $t~(1 \leq t \leq t_h)$ is obtained by:
\begin{equation}
    \begin{aligned}
        \bm{f}_e^t &= \left({f_e^t}_0, ..., {f_e^t}_i, ..., {f_e^t}_{d-1}\right) \in \mathbb{R}^{d}, \\
        \mbox{where}~{f_e^t}_i &= \left\{\begin{aligned}
            &\sin \left(t / 10000^{d/i}\right),     &i \mbox{ is even};\\
            &\cos \left(t / 10000^{d/(i-1)}\right),     &i \mbox{ is odd}.
        \end{aligned}\right. 
    \end{aligned}
\end{equation}
Then, we have the positional coding matrix $f_e$ that describes $t_h$ steps of sequences:
\begin{equation}
    \bm{f}_e = (\bm{f}_e^1, \bm{f}_e^2, ..., \bm{f}_e^{t_h})^T \in \mathbb{R}^{{t_h}\times d}.
\end{equation}
The final Transformer input $X_T$ is the addition of the original input $X$ and the positional coding matrix $f_e$.
Formally,
\begin{equation}
    \bm{X}_T = \bm{X} + \bm{f}_e \in \mathbb{R}^{t_h \times d}.
\end{equation}

\subsection{Transformer Details}

We employ $L = 4$ layers of encoder-decoder structure with $H = 8$ attention heads in each Transformer-based sub-networks.
The MLP$_e$ and the MLP$_d$ have the same shape.
Both of them consist of two fully connected layers.
The first layer has 512 output units with the ReLU activation, and the second layer has 128 but does not use any activations.
The output dimensions of fully connected layers used in multi-head attention layers are set to $d$ = 128.  

\section{Interaction Representation}
\label{Appendix_B}

\begin{figure}[tbp]
\centering
\includegraphics[scale=0.28]{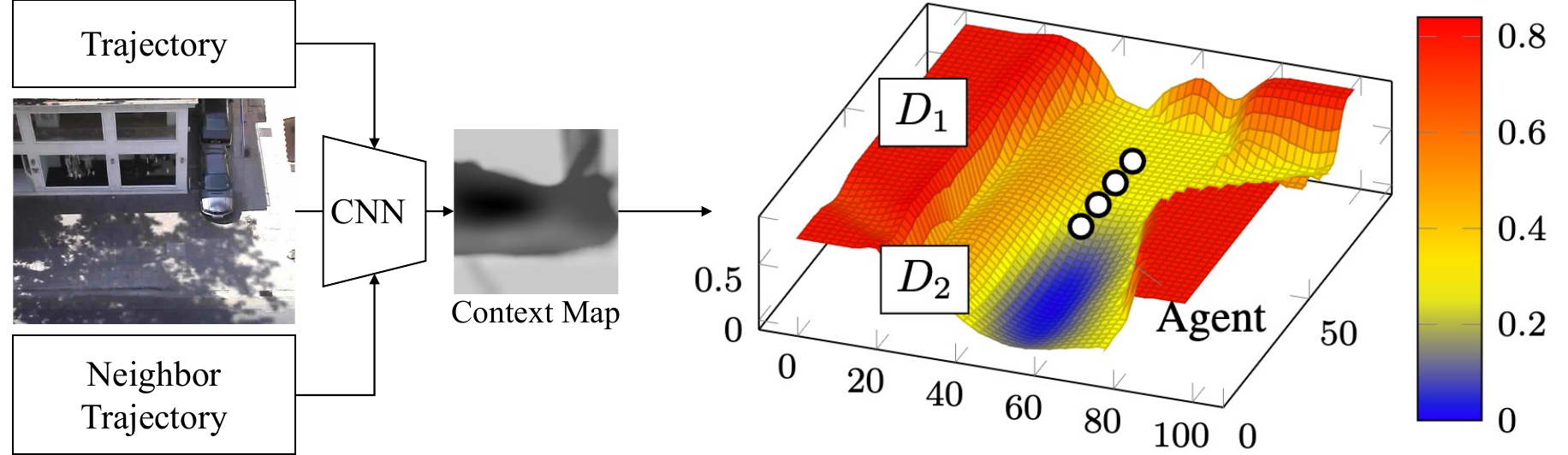}

\caption{
    Visualization of someone's context map.
    A lower semantic label (colored in blue) means that the place has a higher possibility for that agent to move towards, while a higher label (colored in red) means lower possibility.
}
\label{fig_contextmap}
\end{figure}

In the proposed \MODEL, we use the context map \cite{xia2022cscnet} to describe agents' interaction details through a two-dimensional energy map, which is inferred from scene images and their neighbors' trajectories.
Considering both social and scene interactions, it provides potential attraction or repulsion areas for the target agent.
Although the focus of this manuscript is not on the modeling of interactive behaviors, we still show the effect of this part, for it is a crucial research part of the trajectory prediction task.
We visualize one agent's context map in zara1 dataset in \FIG{fig_contextmap}.
The target moves from about $({p_x}_0, {p_y}_0) = (50, 80)$ to the current $({p_x}_1, {p_y}_1) = (50, 50)$ during observation period.

\begin{itemize}
    \item \textbf{Scene constraints}:
        The scene's physical constraints indicate where they could not move.
        The context map gives a higher enough value ($\approx 1$) to show these areas.
        For example, the higher semantic label in the area $D_1 = \{({p_x}, {p_y})| {p_x} \leq 20\}$ reminds pedestrians not to enter the road at the bottom of the zara1 scene.
        Similarly, another high-value area $\{({p_x}, {p_y}) |{p_x} \geq 80, {p_y} \leq 50\}$  reminds pedestrians not to enter the ZARA shop building except the door.
        It illustrates the ability of the context map to model the scene's physical constraints.
    \item \textbf{Social interaction}:
        Social interaction refers to the interactive behaviors among agents, such as avoiding and following.
        The context map does not describe the interaction behavior directly but provides lower semantic labels to areas conducive to agents' passage and higher semantic labels that are not.
        For example, the high-value area $D_2 = \{({p_x}, {p_y})|20 \leq {p_x} \leq 40, {p_y} \leq 80\}$ shows another group of agents' possible walking place in the future who walk towards the target.
        The target agent will naturally avoid this area when planning his future activities.
        Context maps follow the lowest semantic label strategy to describe agents' behaviors.
        A place with a lower semantic label means that the target agent is more likely to pass through.
        Thus, it could show agents' social behaviors in the 2D map directly.
\end{itemize}

\section{Classification Strategy}
\label{Appendix_C}

In the proposed \MODEL, we need to first determine whether different trajectories belong to the same behavioral style mainly based on agents' end-points (or called destinations).
We do not explain the rationale for this classification approach due to the space limitation of the manuscript.
In this section, we will further explore the rationale for using end-points as their behavioral style classification, as well as further explanations of behavioral style classification strategies.

The classification method that we used in \EQUA{eq_alpha_loss1} and \EQUA{eq_alpha_class} can be written together as:
\begin{equation}
    \label{eq_appendix_1}
    \begin{aligned}
        \mbox{Category}(\bm{d}|\mathcal{D}) &= c_s = s, \\
        \mbox{where}~~s &= \underset{i = 1, 2, ..., K_c}{\arg\min} \Vert \bm{D}_i - \bm{d} \Vert.
    \end{aligned}
\end{equation}
The end-points are used as a very important basis for the classification of behavioral styles.
However, in this case, there may be situations where different trajectories have completely different directions although they have the same end-points.

As we mentioned in this manuscript, agents' behavioral preferences tend to be continuously distributed, and it is also more difficult to directly classify these preferences.
The main concern that the manuscript wishes to explore is the multi-behavioral style property of the agent.
With this one constraint, we should fully ensure that the model preserves the different behavioral style features in the trajectory.
Moreover, often more constraints mean fewer possibilities.
Therefore, we want to determine each category by a minimal classification criterion so that each category can ``cover'' more trajectories.

In contrast, if a more strict category judgment approach is used, then each category will cover fewer trajectories, thus requiring a larger number of categories (\IE, a larger $K_c$) to be set to capture more differentiated behavioral preferences.
On the one hand, this has higher data requirements and may make the network difficult to train.
On the other hand, applying the prediction model to more specific scenarios is difficult because the trajectory preferences and interaction behaviors vary significantly in different scenarios.
The strict categorization restriction will also lead to a decrease in the generalization ability of the model to more prediction scenarios.

In addition, we have further investigated and explored the problem of classification strategies.
Specifically, we investigate the trajectory prediction style of the network through a more rigorous classification strategy.
In detail, we expand the end-point $\bm{d}$ into the trajectory $\bm{j}$, and the destinations $\{\bm{D}_i\}_i$ in the set of 2-tuples $\mathcal{D} = \{(\bm{D}_i, c_i)\}_i$ into trajectory keypoints' proposals $\{{\bm{D}_{key}}_i\}_i$ in the new set of 2-tuples $\mathcal{D}_{key} = \{({\bm{D}_{key}}_i, c_i)\}_i$.
Given a set of indexes for the temporal keypoints
\begin{equation}
    \mathcal{T}_{key} = \left\{t_{key}^{1}, t_{key}^{2}, ..., t_{key}^{N_{key}}\right\},
\end{equation}
we have the expanded classification function:
\begin{equation}
    \label{eq_appendix_2}
    \begin{aligned}
        \mbox{Category}(\bm{j}|\mathcal{D}_{key}) &= c_s = s, \\
        \mbox{where}~~s &= \underset{i = 1, 2, ..., K_c}{\arg\min} \sum_{t \in \mathcal{T}_{key}} \Vert {{\bm{D}_{key}}_i}_t - \bm{j}_t \Vert.
    \end{aligned}
\end{equation}
When the set of temporal keypoints contains only the last moment of the prediction period (\IE, $\mathcal{T}_{key} = \{t_h + t_f\}$), the above \EQUA{eq_appendix_2} will degenerate to \EQUA{eq_appendix_1}.

\begin{figure}[tbp]
    \centering
    \includegraphics[width=1.0\linewidth]{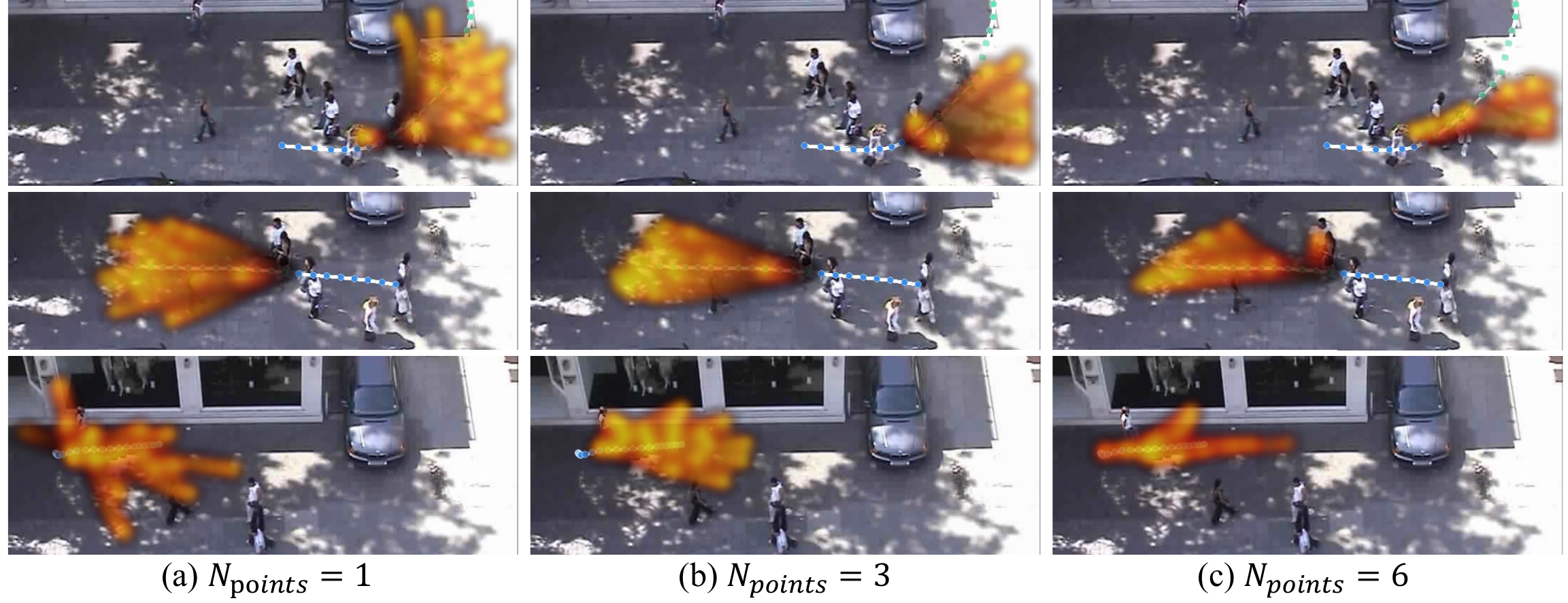}
    \caption{
        The number of trajectory points and prediction styles.
        We show the visualized predictions with different $N_{points}$ configurations.
    }
    \label{fig_appendix_keypoints}
\end{figure}

As shown in Fig. \ref{fig_appendix_keypoints}, we select 1, 3, and 6 keypoints (including the end-points) instead of the classification strategy used like \EQUA{eq_appendix_1}, the network exhibits a completely different prediction style.
The prediction results in the figure show that when there are more constraints on the classification (\IE, 6 keypoints), the predicted results will appear more cautious and lack possibilities and multi-style properties.
Considering that the primary concern of this manuscript is still the multi-style prediction, we choose only one end-point (destination) as the reference for classification.

\end{document}